\documentclass[11pt]{amsart}

\usepackage[
            CJKbookmarks=true,
            bookmarksnumbered=true,
            bookmarksopen=true,
            colorlinks=true,
            citecolor=red,
            linkcolor=blue,
            anchorcolor=red,
            urlcolor=blue,
            pdfauthor={Marius Junge and Kiryung Lee},
            pdfstartview=FitH,
            ]{hyperref}

\usepackage{amsmath,amsxtra,amssymb,amsthm,amsfonts}
\usepackage[usenames]{color}
\usepackage{umoline}
\usepackage[all]{xy}
\usepackage{bm}
\usepackage{bbm}
\usepackage{graphicx}
\usepackage{multirow}

\newtheorem{lemma}{Lemma}[section]

\newtheorem{prop}[lemma]{Proposition}
\newtheorem{theorem}[lemma]{Theorem}
\newtheorem{cor}[lemma]{Corollary}

\newtheorem{rem}[lemma]{Remark}

\newcommand{\re}{\begin{rem}\rm}
  \newcommand{\mar}{\end{rem}}
\newtheorem{exam}[lemma]{Example}

\newtheorem{defi}[lemma]{Definition}

\newcommand{\fo}{\begin{eqnarray*}}
\newcommand{\mel}{\end{eqnarray*}}

\newcommand{\kl}{\pl \le \pl}
\newcommand{\gl}{\pl \ge \pl}

\newcommand{\lel}{\pl = \pl}

\newcommand{\ez}{{\mathbb E}}
\newcommand{\nz}{{\rm  I\! N}}

\newcommand{\rz}{{\mathbb R}}
\newcommand{\zz}{{\mathbb Z}}

\newcommand{\cz}{{\mathbb C}}

\newcommand{\ten}{\otimes}

\newcommand{\pl}{\hspace{.1cm}}

\newcommand{\al}{\alpha}
\newcommand{\si}{\sigma}

\newcommand{\la}{\lambda}
\newcommand{\eps}{\varepsilon}

\newcommand{\E}{{\mathcal E}}

\newcommand{\T}{{\mathcal T}}

\DeclareMathOperator{\absc}{absconv}
\DeclareMathOperator{\Sh}{Sh}

\newcommand{\qd}{\end{proof}\vspace{0.5ex}}

\newcommand{\xspace}{\hbox{\kern-2.5pt}}

\begin{document}

\title[Generalized notions of sparsity and RIP. Part II: Applications]{Generalized notions of sparsity and restricted isometry property. Part II: Applications}

\begin{abstract}
  The restricted isometry property (RIP) is a universal tool for data recovery. We explore the implication of the RIP in the framework of generalized sparsity and group measurements introduced in the Part I paper \cite{junge2016ripI}. It turns out that for a given measurement instrument the number of measurements for RIP can be improved by optimizing over families of Banach spaces. Second, we investigate the preservation of difference of two sparse vectors, which is not trivial in generalized models. Third, we extend the RIP of partial Fourier measurements at optimal scaling of number of measurements with random sign to far more general group structured measurements. Lastly, we also obtain RIP in infinite dimension in the context of Fourier measurement concepts with sparsity naturally replaced by smoothness assumptions.

\end{abstract}

\author{Marius Junge and Kiryung Lee}
\maketitle

\section{Introduction}
\label{sec:intro}

The {\em restricted isometry property} (RIP) has been used as a universal tool in data recovery. In a companion Part I paper \cite{junge2016ripI}, we introduced the generalized notion of sparsity and provided a far reaching generalization of the RIP theory by Rudelson and Vershynin \cite{rudelson2008sparse} and subsequent improvements \cite{rauhut2010compressive,dirksen2015tail} in a unified framework.
In this paper we explore how the RIP results on generalized sparsity models in the Part I paper \cite{junge2016ripI} apply to challenging scenarios not covered by existing theory. Specifically, we illustrate our findings with the examples below.

\subsection{Optimizing RIP with families of Banach spaces} $\atop$

The first example considers the RIP for the canonical sparsity model in $\rz^N$, which is determined by counting the number of nonzero elements. Here measurements are obtained as inner products with functionals $\eta_{kl}$ for $(k,l) \in \zz_N \times \zz_N$, which are given as the translates of the quantum Fourier transform of a fixed measurement instrument $\eta \in \rz^N$ expressed by
\[ \eta_{kl}(j) \lel e^{\frac{2\pi \mathfrak{i} lj}{N}}\eta_{j-k} \pl .\]
The special case of $\eta = [1,\dots,1]^\top \in \rz^N$ has been well studied as partial Fourier measurements (see e.g. \cite{candes2006near,rudelson2008sparse}). We are interested in a scenario where each spectral measurement is taken with a finitely supported window having a specific decaying pattern. The following theorem, obtained as a consequence of the main results in the part I paper \cite{junge2016ripI}, shows how the number of measurements for the RIP can be optimized over a choice of Banach spaces.

\begin{theorem}\label{2} Let $2 < q' \leq \infty$. Suppose that $\|\eta\|_2=\sqrt{N}$.
Then
\[
\sup_{\|x\|_0 \leq r} \Big|\frac{1}{m} \sum_{i=1}^m |(\eta_{k_i,l_i},x)|^2 -\|x\|_2^2\Big| \kl \delta \|x\|_2^2
\]
holds with high probability for independently chosen random pairs $(k_i,l_i)$ provided
\begin{equation}
\label{nummeas}
m \geq c \delta^{-2} (1+\ln m)^3 (q')^3 r^{1-2/q'} \|\eta\|_{q'}^2 \pl.
\end{equation}
By optimizing $m$ in \eqref{nummeas} over $q'$, one obtains the assertion if
\[
m \geq c \delta^{-2} (1+\ln m)^3 {\rm sp}_\eta(r) \pl ,
\]
where the sparsity parameter ${\rm sp}_\eta(r)$ is defined by
\[ {\rm sp}_\eta(r) := \inf_{2 < q' \le \infty} (q')^3 r^{1-2/q'} \|\eta\|_{q'}^2 \pl .\]
\end{theorem}

For the flat vector $\eta = [1,\dots,1]^\top \in \rz^N$ the choice $q'=\infty$ appears optimal. However, for a short window $\eta$ such that the magnitudes show a polynomial decay of order $\al$ for some $\al<1/2$, the optimal choice is given by $q'=1/\al$. Our main tool in this analysis is a flexible framework that derives the RIP for various sparsity models defined by a family of Banach spaces \cite{junge2016ripI}. We recall that $x\in \rz^N$ is $(K,r)$-sparse if
\[
\|x\|_X \kl \sqrt{r} \|x\|_2 \pl ,
\]
where $X$ is a Banach space with unit ball $K$.
For $X \lel \ell_1^N$, we see that $r$-sparse vectors are $(B_1^N,k)$-sparse and more generally $(B_q^N,r^{2/q-1})$-sparse for $1\le q< 2$, where $B_q^N$ denotes the unit ball of $\ell_q^N$. We derive Theorem \ref{2} with $X = \ell_q^N$, where $(q,q')$ is a conjugate pair such that $1/q + 1/q' = 1$.

\subsection{Preserving distance of sparse vectors in generalized models} $\atop$

The conventional notion of sparsity is given by the geometry of a union of subspaces and provides a special feature that the sparsity level is sub-additive. Unfortunately, this property does not hold for our modified sparsity model, which is given by a nonconvex cone. Particularly, compared to the conventional sparsity model, a central drawback in the generalization is that the difference $x-y$ of two $(K,s)$-sparse vectors $x$ and $y$ is no longer $(K,2s)$-sparse. In fact, the adversarial instance of $x-y$ can attain the maximum (trivial) sparsity level. Therefore the RIP does not necessarily imply that the distance of sparse vectors is preserved and one may not distinguish two generalized sparse vectors from their images in low dimension. Instead, we provide a weaker substitute for the generalized model that allows to preserve the distance in certain sense.

Here we adopt the group action arguments to generate measurements \cite{junge2016ripI}. Let $G$ be a finite group with an affine isotropic representation $\si:G \to O_N$ (see Section~\ref{sec:sketching} for a precise definition, but we may work with the transformations given by shifts and modulations as above).

\begin{theorem}
\label{thm:diff}
Let $G$ and $\si$ be as above. Let $g_1,\dots,g_m$ be independent copies of a Haar-distributed random variable on $G$. Let $\eta\in \rz^N$ with $\|\eta\|_2=\sqrt{N}$.
Suppose that
\[
m \gl c \delta^{-2}s(1+\ln m)^3(1+\ln \|\mathrm{Id}:\ell_2^n\to X\|)\|\eta\|_{X^*}^2 \pl .
\]
Then for all $\epsilon > 0$
\[
\Big|\frac{1}{m} \sum_{j=1}^m |\langle \eta,\si(g_j)(x-y) \rangle|^2 - \|x-y\|_2^2 \Big|
\leq \min\Big(\frac{\|x-y\|_2^2}{1+\epsilon},~ \sqrt{2} \delta  (\|x\|_2+\|y\|_2) \|x-y\|_2 \Big)
\]
holds with high probability for all $(K,s)$-sparse vectors $x$ and $y$ such that
\[
\|x-y\|_X \leq \frac{\sqrt{s} \|x-y\|_2}{\sqrt{2} (1+\epsilon) \delta} \pl.
\]
Moreover, the following results hold with high probability for all unit-norm and $(K,s)$-sparse vectors $x$ and $y$. If
\[
\frac{1}{m} \sum_{j=1}^m |\langle \eta,\si(g_j)(x-y) \rangle|^2 \geq 32 \delta^2 \pl,
\]
then
\[
\Big(1-\frac{1}{\sqrt{2}}\Big) \frac{1}{m} \sum_{j=1}^m |\langle \eta,\si(g_j)(x-y) \rangle|^2
\le \|x-y\|_2^2
\le \Big(1+\frac{1}{\sqrt{2}}\Big) \frac{1}{m} \sum_{j=1}^m |\langle \eta,\si(g_j)(x-y) \rangle|^2 \pl.
\]
Otherwise, $\|x-y\|_2 \leq 8 \delta$.
\end{theorem}

The first part of Theorem~\ref{thm:diff} shows that two $(K,s)$-sparse vectors may be distinguished from a small number of their measurements if the difference is sparse up to a certain level (much higher than $s$ for small $\delta$). On the other hand, the second part of Theorem~\ref{thm:diff} implies that one can distinguish two sparse unit-norm vectors if the distance between their measurements is larger than a certain threshold. Otherwise, the vectors are contained in a neighborhood of radius $8\delta$. These results are weaker than the analogous versions with the subadditivity of the sparsity level, but they can be still useful in some applications such as locality-sensitive hashing.

\subsection{Improving group structured measurements with further randomization} $\atop$

In the third illustration, we discuss variations of group structured measurements combined with further randomization. The number of group structured measurements for the RIP, derived in the Part I paper \cite{junge2016ripI}, may scale worse compared to the optimal subgaussian case given by Gordon's lemma \cite{gordon1988milman}. In fact, this was the case with the low-rank tensor example. We propose two different ways to improve the RIP results for group structured measurements with more randomness.

The first approach uses the composition of the group structure measurement system followed by a subgaussian matrix, where the second step further compresses the data. An application of Rosenthal's inequality shows that this composition system also achieves the same optimal scaling as one by a pure subgaussian measurement system. In fact, the group structured measurement system has already reduced the data dimension significantly and the subsequent system given as a small subgaussian matrix requires much less computation compared to the case of applying a single large subgaussian matrix.

The second approach, inspired by a recent RIP result by Oymak et al. \cite{oymak2015isometric}, achieves the optimal scaling by preprocessing the data with multiplication with a random sign pattern before applying the group structured measurement system. This composition system is also interpreted as a single group structured measurement system that employs a larger group for the group actions.

We compare the RIP results for the modified group structured measurements in the example of low-rank tensors. Here the Banach space defining a sparsity model is given as the $d$-fold tensor product of $\ell_2^n$ with the largest tensor norm, which is a natural extension of the bilinear sparsity model given by Schatten 1-class.

\subsection{RIP for infinite dimensional sparsity models} $\atop$

Lastly, we illustrate the RIP for infinite dimensional sparsity models, which are motivated by compressive Fourier imaging. Our goal in this example is to construct a sparsity model in the function space without discretization and establish the RIP theory on this model. In Fourier imaging, the measurements $(y_k)_{k \in \zz}$ are obtained as
\[
y_k \lel \int_{-\infty}^\infty \overline{\psi_k(t)} f(t) dt, \quad k \in \zz \pl,
\]
where $(\psi_k)_{k\in\zz}$ denotes the complex sinusoid defined by $\psi_k(t)=e^{2\pi \mathfrak{i}kt}$ for $t \in [0,1)$ and $f$ is a signal in $L_2(0,1)$. When the support of $f$ is restricted to $[0,1)$, the sequence $(y_k)_{k \in \zz}$ corresponds to the Fourier series $(\hat{f}(k))_{k \in \zz}$ and the map from $f$ to $(\hat{f}(k))_{k \in \zz}$ is bijective.


We consider measurements in the form of
\begin{equation}
\label{fourier_meas_ip}
\sum_{k=-N}^{N-1}\widehat{\tau_t f}(k)
= \sum_{k=-N}^{N-1} e^{-2\pi \mathfrak{i}k t} \widehat{f}(k) \pl,
\end{equation}
where $\tau_t$ denotes the circular shift operator. We would like to preserve the $\ell_2$ norm of $(\hat{f}(k))_{-N\le k<N}$ by measurements from $m$ translates of $f$, which are given in \eqref{fourier_meas_ip} for $t = t_1,\dots,t_m$. Ideally, in the noise-free case, $m = 2N$ shifts can provide unique identification of $(\hat{f}(k))_{-N\le k<N}$. However, in this case, the measurement system is ill-conditioned. However, as we show in the following theorem, a sparsity model in $L_2(0,1)$ allows that the $\ell_2$-norm of the subsequence $(\hat{f}(k))_{-N\le k<N}$ is preserved by fewer measurements.

\begin{theorem}
Let $\|\cdot\|_{2,w}$ be a seminorm on $L_2(0,1)$ defined by
\[
\|f\|_{2,w} = \Big( \sum_{k=-N}^{N-1} |\hat{f}(k)|^2 \Big)^{1/2} \pl
\]
and
\[
K_{\rho,\gamma} \lel \{f \in L_2(0,1) \cap C^1 \pl|\pl \|f'\|\kl \rho \|f\|_{L_2},\pl \la({\rm supp}(f))\kl \gamma \}
\]
for $\rho \le N/2$ and $\gamma \in (0,1)$, where $\lambda(\cdot)$ denote the Lebesgue measure. Suppose that $t_1,\dots,t_m$ are independent copies of a uniform random variable on $(0,1)$.
Then $m = O(\delta^{-2} \gamma (N^2 + \rho^2) \ln^7 N)$ suffices to satisfy
\[
\sup_{f \in K_{\rho,\gamma},~ \|f\|_{2,w} = 1} \Big|\frac{1}{m}\sum_{j=1}^m
\Big| \sum_{k=-N}^{N-1} e^{-2\pi \mathfrak{i}k t_j} \hat{f}(k) \Big|^2 -\|f\|_{2,w}^2\Big| \leq \max(\delta,\delta^2)
\]
with high probability.
\end{theorem}

The sparsity model $K_{\rho,\gamma}$ is determined by two parameters $\rho$ and $\gamma$. Sparsity is measured as the relative occupancy $0 < \gamma < 1$ in the Lebesgue measure and smoothness is measured by the other parameter $\rho$. The seminorm $\|f\|_{2,w}$ is preserved by fewer translates for smaller $\gamma$. Particularly, the number of measurements $m$ can be sublinear for $\gamma = o(1/N)$.

Our theory improves on existing results in several ways described below.
First, unlike the conventional compressed Fourier imaging (e.g., \cite{candes2006robust}),
our measurement model does not involve any discretization and is consistent with the physics in acquisition systems. Second, our setup employs a more flexible sparsity model and considers a realistic scenario where only finitely many measurements are available. Sub-Nyquist sampling of multiband signals \cite{feng1996spectrum,feng1998universal,mishali2009blind,mishali2011xampling} is considered as compressed sensing of analog sparse signals. The multiband sparse signal model in $f\in L_2(0,1)$ is defined so that the support is restricted to be on a few active blocks of $[0,1)$. This is far more restrictive than our infinite dimensional sparsity model.

\subsection{Organization}

Thus far we have demonstrated snapshots of our results in their simplified forms. Full results in detail on each topic are presented in later sections as follows. Section~\ref{sec:optrip} discusses how the main results in the Part I paper \cite{junge2016ripI} can be utilized to optimize the number of measurements for the RIP over the choice of Banach spaces and 1-homogeneous functions. This result is illustrated with the example of partial windowed Fourier transform and its noncommutative version. Section~\ref{sec:sketching} provides a general framework that preserves the distance of sparse vectors without the subadditivity of the sparsity level. Section~\ref{sec:composition} proposes two ways to further improve the number of group structured measurements for the RIP with more randomness. Obtained results are compared over the low-rank tensor example. Lastly, in Section~\ref{sec:infdim}, we illustrate how (semi)norms are preserved by finitely many measurements using various infinite dimensional sparsity models and identify the number of measurements in each scenario.

\subsection{Notation}

In this paper, the symbols $c,c_1,c_2,\dots$ and $C,C_1,C_2,\dots$ will be reserved for numerical constants, which might vary from line to line.
We will use notation for various Banach spaces and norms.
The norm of a Banach space $X$ is denoted by $\|\cdot\|_X$.
For example, $\|\cdot\|_{L_q}$ denote the norm that defines $L_q(0,1)$.
We will use the shorthand notation $\|\cdot\|_p$ for the $\ell_p$-norm for $1\leq p\leq \infty$.
The operator norm will be denoted by $\|\cdot\|$.
For $N \in \nz$, the unit ball in $\ell_p^N$ will be denoted by $B_p^N$.
The identity operator will be denoted by $\mathrm{Id}$.

\section{Optimizing Restricted Isometry Property}
\label{sec:optrip}

In this section, we present two examples where one can optimize the number of measurements for the RIP with respect to given conventional sparsity models. Specifically, we consider the canonical sparsity model and low-rank matrix model. In the literature, relaxation of these models with corresponding Banach spaces ($\ell_1^n$ for the canonical sparsity model and $S_1^n$ for the low-rank matrix model) provided the RIP from a near optimal number of incoherent measurements. However, in practice, there exist physical constraints on designing the measurement system and ideally incoherent instruments are not always available. In this situation, we demonstrate that the number of measurements for the RIP can be optimized via our general framework in the Paper I \cite{junge2016ripI}.

\subsection{RIP of subsampled short-time discrete Fourier transform} $\atop$
\label{sec:stdft}

The first example provides the RIP of a partial short-time Fourier transform, which can be considered as a non-ideal version of a partial Fourier operator.
Let $h: \mathbb{Z}_N \to \mathbb{C}$ be a window function.
The windowed discrete Fourier transform of $f : \mathbb{Z}_N \to \mathbb{C}$ is given by
\begin{equation}
\label{wdft}
c(t,k) = \sum_{\ell=0}^{N-1} f(\ell) \overline{h(\ell-t)} e^{-\mathfrak{i}2\pi k\ell/N}, \quad \forall t, k \in \mathbb{Z}_N \pl,
\end{equation}
where the time indices are modulo $N$.
Let $\sigma: \mathbb{Z}_N \times \mathbb{Z}_N \to O_N$, where $O_N$ is the orthogonal group, be an isotropic affine representation given by
\[
\sigma(t,k) = \Lambda^t \mathrm{Sh}^k,
\]
where $\Lambda$ is the usual modulation defined by
\[
\Lambda(\bm{e}_l)=e^{\mathfrak{i} 2\pi l/N} \bm{e}_l, \quad \forall l = 1,\dots,N \pl,
\]
and $\Sh$ is the circular shift modulo $N$ such that
\[
\Sh(\bm{e}_l) =
\begin{cases}
\bm{e}_{l+1} & 1 \leq l \leq N-1 \pl ,\\
\bm{e}_1 & l = N \pl.
\end{cases}
\]
Here, $\bm{e}_1,\dots,\bm{e}_N$ are the standard basis vectors in $\mathbb{R}^N$.
Let $\eta = [h[0],\dots,h[N-1]]^\top$ and $x = [f[0],\dots,f[N-1]]^\top$.
Then the windowed DFT coefficient in \eqref{wdft} is a group measurement given by
\[
c(t,k) = \langle \eta, \sigma(t,k) x \rangle, \quad
\forall t, k \in \mathbb{Z}_N \pl.
\]

In signal processing, particularly for large $N$, it is usual to take a spectral measurement from a finite block of the signal $f$. Then the resulting measurements correspond to short-time discrete Fourier transform (STDFT) of $f$ with a given window function $h$. Typically, to avoid the leakage due to the discontinuity at the boundary, windows are designed with decaying magnitudes.

We consider the RIP of subsampled STDFT measurements on $k$-sparse signals with respect to the canonical sparsity model, i.e. on the set $S = \{x \in \cz^N \pl|\pl \|x\|_0 \leq k\}$, where $\|\cdot\|_0$ counts the number of nonzero elements. In our generalized notion of sparsity, there exist convex sets $K \subset B_2^N$ such that every $k$-sparse vector $x$ in $S$ is $(K,s)$-sparse, i.e. $\|x\|_X \leq \sqrt{s} \|x\|_2$, where $X$ is the Banach space with unit ball $K$. For example, one can choose $X=\ell_1^N$ with $s=k$ but this is not the only choice. By using our general framework from the Part I paper \cite{junge2016ripI}, it is possible to optimize the number of measurements for the RIP on $S$ over the choice of the convex set $K$ and the parameter $s$ as shown in the following theorem.

\begin{theorem}[Partial STDFT with a decaying window]
\label{thm:stft}
Let $(t_1,k_1),\dots,(t_m,k_m)$ be independent copies of a uniform random variable on $\mathbb{Z}_N \times \mathbb{Z}_N$. For $\eta = [\eta_1,\dots,\eta_N]^\top$, let $(\eta_j^\downarrow)_{1\leq j\leq N}$ denotes the rearrangement of $(|\eta_j|)_{1\leq j\leq N}$ in the non-increasing order.
Suppose that
\[
\eta_j^\downarrow =
\begin{cases}
c_{\alpha,N,N_\eta} j^{-\alpha} & j \leq N_\eta \pl ,\\
0 & j < N_\eta \pl .
\end{cases}
\]
for $0<\al<1/2$ and $N_\eta < N$, where $c_{\alpha,N,N_\eta}$ is a constant such that $\|\eta\|_2 = \sqrt{N}$. Then there exists a numerical constant $c$ such that
\begin{equation}
\label{stft:rip}
\mathbb{P}\left(
\sup_{\|x\|_0 \leq k, ~\|x\|_2 = 1}
\Big| \frac{1}{m} \sum_{j=1}^m \Big|\langle \sigma(t_m,k_m) \eta, x \rangle\Big|^2 - \|x\|_2^2 \Big| \geq \delta
\right) \leq \zeta
\end{equation}
provided
\[
m \geq \frac{c \delta^{-2} k \max( \al^{-3} (1+\ln m)^3, \ln(\zeta^{-1}) ) N_\eta^{2\alpha} (1+\ln N_\eta)^{2\alpha} N}{k^{2\alpha} N_\eta(1-N_\eta^{-1+2\alpha})} \pl .
\]
\end{theorem}

\begin{rem}
{\rm
Alternatively, applying \cite[Theorem~5.1]{junge2016ripI} with $X = \ell_1^N$ provides the same RIP as in Theorem~\ref{thm:stft} for
\begin{equation}
\label{stft:rip:classical}
m \geq \frac{c \delta^{-2} k \max( (1+\ln m)^3(1+\ln N),~ \ln(\zeta^{-1}) ) N_\eta^{2\al} N}{N_\eta(1-N_\eta^{-1+2\alpha})} \pl .
\end{equation}
Note that the optimized number of measurements for the RIP in Theorem~\ref{thm:stft} is smaller than that in \eqref{stft:rip:classical} by factor $k^{2\alpha}$.
}
\end{rem}

\begin{proof}[Proof of Theorem~\ref{thm:stft}]
Let $1 \leq q < 2$ and $X = \ell_q^N$.
Then it follows that since $X^* = \ell_{q'}^N$, where $2 < q' < \infty$, is of type 2 and the type 2 constant $T_2(X^*)$ is upper bounded by $\sqrt{q'}$ \cite[Lemma~3]{carl1985inequalities}.
Let $s = k^{1-2/q'}$. Then each $k$-sparse $x$ satisfies $\|x\|_q \leq \sqrt{s} \|x\|_2$.
Therefore, by \cite[Theorem~5.3]{junge2016ripI}, the assertion in \eqref{stft:rip} holds if
\begin{equation}
\label{eq1:pf:stft:rip}
m \geq c \delta^{-2} k^{1-2/q'} \max( (q')^3 (1+\ln m)^3, \ln(\zeta^{-1}) ) \|\eta\|_{X^*} \pl .
\end{equation}

It remains to compute $\|\eta\|_{X^*}$.
First of all, the normalization constant $c_{\alpha,N,N_\eta}$ satisfies
\[
N = \|\eta\|_2^2 = c_{\alpha,N,N_\eta}^2 \sum_{j=1}^{N_\eta} j^{-2\alpha}
\geq c_{\alpha,N,N_\eta}^2 (N_\eta^{1-2\alpha}-1).
\]
Next we compute $k^{1-2/q'} \|\eta\|_{q'}^2$ in the following three cases for $r'$.

\noindent{\bf Case~1:} $q' \alpha < 1$
\begin{align*}
k^{1-2/q'} \|\eta\|_{q'}^2
& = k^{1-2/q'} c_{\alpha,N,N_\eta}^2 \left( \sum_{j=1}^{N_\eta} j^{-q' \alpha} \right)^{2/q'}
\leq \frac{k^{1-2/q'} N N_\eta^{2/q'-2\alpha}}{N_\eta^{1-2\alpha}-1} \\
& = k \left(\frac{N_\eta}{k}\right)^{2/q'} \frac{N}{N_\eta(1-N_\eta^{-1+2\alpha})} \pl . \end{align*}
\noindent{\bf Case~2:} $q' \alpha = 1$
\begin{equation}
\label{eq2:pf:stft:rip}
k^{1-2/q'} \|\eta\|_{q'}^2
\leq \frac{k^{1-2\alpha} N (1 + \ln N_\eta)^{2\alpha}}{N_\eta^{1-2\alpha}-1}
= k \left(\frac{N_\eta}{k}\right)^{2\alpha} \frac{N (1+\ln N_\eta)^{2\alpha}}{N_\eta(1-N_\eta^{-1+2\alpha})} \pl .
\end{equation}
\noindent{\bf Case~3:} $q' \alpha > 1$
\[
k^{1-2/q'} \|\eta\|_{q'}^2
\leq \frac{C k^{1-2/q'} N}{N_\eta^{1-2\alpha}-1}
= C k^{1+2\al-2/q'} \left(\frac{N_\eta}{k}\right)^{2\al} \frac{N}{N_\eta(1-N_\eta^{-1+2\alpha})} \pl .
\]
Note that the second case has the smallest upper bound.
Applying \eqref{eq2:pf:stft:rip} to \eqref{eq1:pf:stft:rip} completes the proof.
\qd

\subsection{Deterministic instrument for Schatten classes} $\atop$

Next, we show an analogous result in the noncommutative case.
Here, we consider the RIP of the group structured measurements restricted to the set of rank-$r$ matrices, where the affine representation corresponds to the double quantum Fourier-transform, i.e.
\[
\si(k,j,k',j')(a) \lel \Lambda^k\Sh^ja(\Sh^{j'})^*\Lambda^{-k'}, \quad \forall a \in M_n \pl.
\]
Let $g$ be a Haar-distributed random variable on $\zz_n^4$. Then
\[
\int_{\zz_n^4} \si(g)T\si(g)^* d\mu(g) \lel \frac{\mathrm{tr}(T)}{n^2}\mathrm{Id}
\]
holds for all linear operator $T$ on $M_n$. Hence the affine representation is isotropic.

We first recall the RIP result for an arbitrary instrument $\eta$ in this case.

\begin{prop} Let $1\le q\le 2$ and $\eta \in S_{q'}^n$ be a fixed vector such that $\|\eta\|_{S_2} \lel n$. Then
\[
\mathbb{P}\left(
\sup_{\|a\|_{S_q} \le \sqrt{s},~ \|a\|_{S_2} = 1}
\Big|\frac{1}{m} \sum_{l=1}^m |\langle\si(g_l)(\eta), a\rangle|^2-\|a\|_{S_2}^2\Big|
\geq \delta
\right) \leq \zeta
\]
holds provided
\[
\frac{m}{(q')^3 (1+\ln m)^3 + \ln(\zeta^{-1})} \gl c \delta^{-2}s \|\eta\|_{S_{q'}}^2 \pl .
\]
\end{prop}

\begin{proof} The Schatten class $S_{q'}$ has type $2$ with constant $c_1 \sqrt{q'}$ \cite{tomczak1989banach}. Therefore the main technical result in \cite[Theorem~5.3]{junge2016ripI} applies here with $p=2$. \qd

Moreover, for a rank-$r$ matrix we always have
\[
\|x\|_{S_q} \kl r^{1/q-1/2} \|x\|_{S_2}
\]
and hence may apply the previous result for $s\lel r^{2/q-1}$ and get the following corollary.

\begin{cor}
\label{cor:ncopt}
Let $\eta \in M_n$ be a fixed vector such that $\|\eta\|_{S_2} \lel n$. Then
\[
\mathbb{P}\left(
\sup_{{\rm rk}(a)\le r,~ \|a\|_{S_2}=1}
\Big|\frac{1}{m} \sum_{l=1}^m |\langle\si(g_l)(\eta), a\rangle|^2-\|a\|_{S_2}^2\Big|
\geq \delta
\right) \leq \zeta
\]
holds provided
\[ m \gl c \delta^{-2} \mathrm{sp}_\eta(r) \Big( (1+\ln m)^3 + \ln(\zeta^{-1}) \Big) \pl ,\]
where
\[
\mathrm{sp}_\eta(r) := \inf_{q'\ge2} r^{1-2/q'} (q')^3 \|\eta\|_{S_{q'}}^2 \pl .
\]
\end{cor}

The parameter $\mathrm{sp}_\eta(r)$ in Corollary~\ref{cor:ncopt} denotes the optimized sparsity level, which is determined by the rank $r$ and the instrument $\eta$.
Next we demonstrate the number of group structured measurements for the RIP given by Corollary~\ref{cor:ncopt} for particular choices of the instrument $\eta$.
The first example considers an ideal instrument that generates incoherent measurements.

\begin{exam}
Let $\eta =\sqrt{n} \bm{1}_{M_n}$, where $\bm{1}_{M_n}$ is the $n$-by-$n$ identity matrix. Then $\|\xi\|_{S_2}=n$. For this particular choice of $\eta$, we also have
\[
\si(k,j,k',j')(\eta)
\lel \Lambda^{k}\Sh^{j-j'}\Lambda^{-k'} \eta
\lel e^{\frac{\mathfrak{i} 2\pi(j-j')k'}{n}}\Lambda^{k-k'}\Sh^{j-j'} \eta \pl .
\]
Therefore, the number of distinct elements in the orbit of $\eta$ is $n^2$ instead of $n^4$. In other words, we sample from $n^2$ possible measurements.
On the other hand, since $\|\xi\|_{S_{q'}}^2 \lel n^{1+2/q'}$, choosing $q' = 1+\ln n$ gives $\mathrm{sp}_\eta(r) = O(r n (1+\ln n)^3)$. The factor $n$ accounts for geometry of the Schatten class and will disappear in the commutative case.
\end{exam}

Similar to the partial STFT example in the commutative case, the second example considers a non-ideal instrument with fast decaying singular values.

\begin{exam}
Let $\eta$ satisfy $\|\eta\|_{S_2} = n$ and
\[ s_j(\eta) \lel c_{\al,n} j^{-\al}, \quad \forall j=1,\dots,n \]
for $\al < 1/2$, where $s_j(\eta)$ denotes the $j$th singular value of $\eta$ in the non-increasing order.
Similar to the proof of Theorem~\ref{thm:stft}, we show
\[
r^{1-2/q'} (q')^3 \|\eta\|_{S_q'}^2
\le
\begin{cases}
r n (q')^3 \left(\frac{n}{r}\right)^{2/q'} \frac{1}{1-n^{-1+2\alpha}} & q' \alpha < 1 \pl ,\\
r n (q')^3 \left(\frac{n}{r}\right)^{2\alpha} \frac{(1+\ln n)^{2\alpha}}{1-n^{-1+2\alpha}} & q' \alpha = 1 \pl ,\\
C r n (q')^3 r^{2\al-2/q'} \left(\frac{n}{k}\right)^{2\al} \frac{1}{1-n^{-1+2\alpha}} & q' \alpha > 1 \pl .
\end{cases}
\]
We see that the function $(n/r)^{2/q'}$ is decreasing for $q'<\al^{-1}$ and the function $r^{2\al-2/q'}$ is increasing for $q'>\al^{-1}$. Thus $q=\al^{-1}$ is the best choice and we
deduce that ${\rm sp}_\eta(r) = O(\al^{-3} r n (n/r)^{2\al} (1+\ln n)^{2\al})$.
This upper bound on the number of measurements for RIP is smaller than the choice of $q'=1+\ln n$. Unlike the previous example, the instrument produces less incoherent measurements, which can be compensated with a penalty in the number of measurements for small $r$.
\end{exam}

\section{RIP on difference of sparse vectors}
\label{sec:sketching}

In the conventional sparsity models given by a union of subspaces, the sparsity level measured by either the number of nonzeros or the rank satisfies the {\em sub-additivity}. An important implication is that the difference of two sparse vectors is still sparse up to the sum of the sparsity levels of each sparse vector. However, this is not the case for generalized sparsity models given by a nonconvex cone. Therefore, the RIP does not automatically preserve the distance between sparse vectors in the general setup. However, with some careful arguments, one can show that the distance is still preserved in some weaker sense. In this section, we will discuss this problem using the notion of the multiresolution restricted isometry property (MRIP).

The MRIP was originally proposed for the canonical sparsity model by Oymak et al. \cite{oymak2015isometric}. We generalize it to general sparsity models with a slight modification. Let $H$ be a Hilbert space and $X$ be a Banach space with unit ball $K \subset B_H$, where $B_H$ denotes the unit ball in $H$. Note that if $K$ has a non-empty interior there exists a number
\begin{equation}
\label{smax}
s_{\max}(K) = \|\mathrm{Id}:X\to H\|^{-1/2}
\end{equation}
such that $\|x\|_X\le \sqrt{s_{\max}(K)} \|x\|_H$ holds for all $x$. For example, if $H=\ell_2^N$ and $X=\ell_1^N$, then $s_{\max}(K) = N$. Given the definition of $s_{\max}(K)$, we state the definition of the MRIP as follows.

\begin{defi}[Multiresolution restricted isometry property]
\label{mrip_gen}
Let $H$ be a Hilbert space, $K \subset H$ be a convex set, $X$ be a Banach space with unit ball $K$, and $s_{\max}(K)$ be a constant defined in \eqref{smax}.
We say that $A:H\to\ell_2^m$ satisfies the MRIP with distortion $\delta > 0$ at sparsity level $s \geq 1$ if
\[
\sup_{\|x\|_X \leq \sqrt{2^l s},~ \|x\|_H = 1} |\|Ax\|_2^2 - \|x\|_H^2| \leq 2^{l/2} \max(2^{l/2} \delta,~ 2^l \delta^2) \pl
\]
holds for all $\lfloor -\log_2 s \rfloor \leq l \leq \lceil \log_2 (s_{\max}(K)/s) \rceil$.
\end{defi}

The following lemma shows that the MRIP can preserve the distance of two sparse vectors when the sparsity level of the difference is below a certain threshold.

\begin{lemma}\label{sketch}
Let $H$, $K$, $X$, and $s_{\max}(K)$ be defined as above,
$\delta>0$, and $s \in \nz$. Suppose that $A:H\to\ell_2^m$ satisfies the MRIP with distortion $\delta$ at sparsity level $s$. Then for all $x,y\in H$
\begin{align*}
|
\|A x - A y\|_2^2 - \|x-y\|_H^2
|
\le
\max\Big(
\frac{\sqrt{2}\delta \|x-y\|_X \|x-y\|_H}{\sqrt{s}},~
\frac{2\delta^2 \|x-y\|_X^2}{s}
\Big) \pl.
\end{align*}
Moreover for any $\epsilon > 0$
\begin{equation}
\label{sketch:bnd0}
|
\|A x - A y\|_2^2 - \|x-y\|_H^2
|
\leq \min\Big(\frac{\|x-y\|_H^2}{1+\epsilon},~ \sqrt{2} \delta  (\|x\|_H+\|y\|_H) \|x-y\|_H \Big)
\end{equation}
provided that $x,y$ are $(K,s)$-sparse and
\[
\|x-y\|_X \leq \frac{\sqrt{s} \|x-y\|_H}{\sqrt{2} (1+\epsilon) \delta} \pl.
\]
\end{lemma}

\begin{rem}
{\rm
Lemma~\ref{sketch} preserves the distance of two $(K,s)$-sparse vectors $x,y$ by \eqref{sketch:bnd0} if the sparsity level of $x-y$ is below the threshold $s/2(1+\epsilon)\delta^2$, which is higher than the sparsity levels $s$ of $x$ and $y$ for small $\delta$. The estimate in \eqref{sketch:bnd0} implies that the distortion is strictly less than $\|x-y\|_H^2$, which implies a local injectivity.

Since the Hilbert space norm $\|\cdot\|_H$ is preserved by the RIP up to a small distortion $\delta$, we can always compare two spare vectors after normalization. Suppose that $\|x\|_H = \|y\|_H = 1$. Then \eqref{sketch:bnd0} also implies that the distortion is no bigger than $2 \sqrt{2} \delta \|x-y\|_H$. Although this distortion bound is more conservative than $\delta \|x-y\|_H^2$, which is available if the sparsity level is subadditive, it can be still useful for certain applications. For example, similar deviation bounds have been used in the analysis of iterative optimization algorithms for matrix completion (see \cite[Lemma~5]{chen2015fast} and \cite[Lemma~8]{zheng2016convergence}). We expect that this weak preservation of the difference of two sparse vectors can be useful in generalizing existing theory to a wider class of sparsity model.
}
\end{rem}

\begin{proof}[Proof of Lemma~\ref{sketch}]
Let $h\lel x-y$ denote the difference between $x$ and $y$.
We may find $l$ such that $\lfloor -\log_2 s \rfloor \le l \le \lceil \log_2 (s_{\max}(K)/s) \rceil$ and
\[
2^l s < \frac{\|h\|_X^2}{\|h\|_H^2} \kl 2^{l+1} s \pl .
\]
Then $h$ is $s_{l+1}$-sparse, where $s_{l+1} = 2^l s$, and hence
\[
|
\|A h\|_2^2
-\|h\|_H^2
|
\kl \max\{\delta_{l+1},\delta_{l+1}^2\} \|h\|_H^2 \pl ,
\]
where $\delta_{l+1} = 2^{(l+1)/2}\delta$.
Let us first assume that $\delta_{l+1} < 1$. Then we get
\begin{equation}
\label{sketch:est1}
\delta_{l+1} \|h\|_H^2 = 2^{(l+1)/2}\delta \|h\|_H^2
\le \sqrt{2} \delta s^{-1/2} \|h\|_X \|h\|_H \pl.
\end{equation}
In case $\delta_{l+1} \ge 1$ we get
\begin{equation}
\label{sketch:est2}
\delta_{l+1}^2 \|h\|_H^2 = 2^{l+1} \delta^2 \|h\|_H^2
\le 2 \delta^2 s^{-1} \|h\|_X^2 \pl .
\end{equation}
This proves the first assertion.
Next, for the second assertion, we assume that $x,y$ are $(K,s)$-sparse. Then we have
\begin{equation}
\label{sketch:bnd}
\|h\|_X \kl \|x\|_X+\|y\|_X\kl \sqrt{s} (\|x\|_H+\|y\|_H) \pl .
\end{equation}
If $h = x-y$ additionally satisfies
\[
\|h\|_X \leq \frac{\sqrt{s}}{\sqrt{2}(1+\epsilon)\delta} \|h\|_H,
\]
then
\[
2^l s < \frac{\|h\|_X^2}{\|h\|_H^2} \le \frac{s}{2(1+\epsilon)^2\delta^2} \pl ,
\]
which implies $\delta_{l+1} = 2^{(l+1)/2} \delta \leq \frac{1}{1+\epsilon} < 1$.
Therefore, we apply the estimate in \eqref{sketch:est1} with the upper bound in \eqref{sketch:bnd}.
\qd



The following corollary of Lemma~\ref{sketch} shows that two unit-norm and $(K,s)$-sparse vectors can be distinguished if they are well separated in the measurement domain. Otherwise, $x$ and $y$ are close in the original space. This result is weaker than the preservation of the distance but applies to a wider class of models. This result can be used in some applications such as locality-sensitive hashing.

\begin{cor}
\label{cor:weak_diff}
Suppose the hypothesis of Lemma \ref{sketch}. Let $x,y$ be unit-norm and $(K,s)$-sparse vectors. If
\begin{equation}
\label{eq:largedist}
\|A x - A y\|_2 \geq 4\sqrt{2} \delta,
\end{equation}
then
\begin{equation}
\label{eq:weak_diff_bnd}
\Big(1-\frac{1}{\sqrt{2}}\Big) \|Ax-Ay\|_2^2
\le \|x-y\|_H^2
\le \Big(1+\frac{1}{\sqrt{2}}\Big) \|Ax-Ay\|_2^2.
\end{equation}
Otherwise,
\[
\|x-y\|_H \leq 8 \delta.
\]
\end{cor}

\begin{proof} Let $h=x-y$. Then by Lemma~\ref{sketch}, we have
\begin{equation}
\label{eq:res_sketch}
\begin{aligned}
|\|A h\|_2^2 - \|h\|_H^2|
&\le \max\Big(
\frac{\sqrt{2}\delta \|x-y\|_X \|x-y\|_H}{\sqrt{s}},~
\frac{2\delta^2 \|x-y\|_X^2}{s}
\Big)\\
&\le \max\{2\sqrt{2}\delta \|h\|_H,8\delta^2\} \pl .
\end{aligned}
\end{equation}

We proceed the proof in the following two complementary cases.

\noindent\textbf{Case 1:} $\|h\|_H \geq \|A h\|_2$. \\
It follows that $2\sqrt{2} \delta \|h\|_H \geq 16 \delta^2$ and the maximum in \eqref{eq:res_sketch} is attained in the first term. Thus by \eqref{eq:largedist} and \eqref{eq:res_sketch}, we have
\[
\|h\|_H^2 - \|A h\|_2^2 \leq 2\sqrt{2} \delta \|h\|_H \leq \frac{\|h\|_H^2}{2},
\]
which implies $\|h\|_H \leq \sqrt{2} \|A h\|_2$.
Then by \eqref{eq:largedist} and \eqref{eq:res_sketch} we have
\[
|\|A h\|_2^2 - \|h\|_H^2| \leq 4 \delta \|A h\|_2 \leq \frac{\|A h\|_2^2}{\sqrt{2}},
\]
which implies \eqref{eq:weak_diff_bnd}.

\noindent\textbf{Case 2:} $\|h\|_H \leq \|A h\|_2$. \\
By \eqref{eq:res_sketch} and the upper estimate for $\delta\le \frac{\|Ah\|}{4\sqrt{2}}\delta$ we deduce
\[
|\|A h\|_2^2 - \|h\|_H^2| \leq 2\sqrt{2} \delta \|A h\|_2 \leq \frac{\|A h\|_2^2}{2},
\]
which implies \eqref{eq:weak_diff_bnd}.
Thus the first part is proved.

For the second part,
we suppose that $\|A h\|_2 < 4\sqrt{2} \delta$ and $\|h\|_H > 8 \delta$ hold simultaneously.
By \eqref{eq:res_sketch} and the fact that the first term achieves the maximum in in \eqref{eq:res_sketch},
\[
\|h\|_H^2 - \|A h\|_2^2 \leq 2\sqrt{2} \delta \|h\|_H \leq \frac{\|h\|_H^2}{2\sqrt{2}}.
\]
Then it follows that
\[
\|h\|_H^2 \leq \frac{2\sqrt{2}}{2\sqrt{2}-1} \|A h\|_2^2 < 64 \delta^2 < \|h\|_H^2,
\]
which is a contradiction. Therefore, $\|A h\|_2 < 4\sqrt{2} \delta$ implies $\|h\|_H \leq 8 \delta$. This completes the proof.\qd

\begin{rem} {\rm We did not optimize the constants in Corollary~\ref{cor:weak_diff}. More generally, we find a conditional RIP property below. Fix $\alpha > 2\sqrt{2}$. Let $\beta > 0$ satisfy $\alpha^2 < \beta(\beta- 2\sqrt{2})$.
If $\|Ah\|_2 \gl \alpha \delta$, then
\[
\left(1-\frac{2\sqrt{2}}{\sqrt{\alpha(\alpha-2\sqrt{2})}}\right) \|Ah\|_2^2
\kl \|h\|_H^2
\kl \left(1+\frac{2\sqrt{2}}{\sqrt{\alpha(\alpha-2\sqrt{2})}}\right)\|Ah\|_2^2 \pl .
\]
Otherwise, $\|h\|_H \kl \beta \delta$. One can optimize the constants $\alpha$ and $\beta$ to tighten the estimate.
} \end{rem}

\begin{rem}
{\rm Corollary~\ref{cor:weak_diff} implies that if the distance between $Ax$ and $Ay$ for two unit-norm sparse vectors $x$ and $y$ is larger than $4\sqrt{2} \delta$, then the distance between $x$ and $y$ in $H$ is equivalent to $\|Ax - Ay\|_2$ up to a constant factor.
In other words, one can distinguish $x$ and $y$ from their linear measurements.
However, if $Ax$ and $Ay$ are close by satisfying $\|Ax - Ay\|_2 < 4\sqrt{2} \delta$, then Corollary~\ref{cor:weak_diff} only confirms that $\|x-y\|_H$ is less than $8\delta$, i.e. one cannot distinguish two similar sparse vectors $x$ and $y$ from their measurements.
Note that we did not optimize the constants in Corollary~\ref{cor:weak_diff}.
Obviously, this result is weaker than the uniform preservation of distance of any two sparse vectors (regardless of the amount of distance) given by RIP with respect to an exact sparsity model. However, this weak property will be still useful in applications. For example, in clustering sparse vectors, if the centroids of clusters are well separated via the dimensionality reduction via $A$, then one can compute clustering in the compressed domain.
}
\end{rem}

\begin{rem}
{\rm
Corollary~\ref{cor:weak_diff} provides a recovery guarantee by a nonconvex programming.
Suppose that $x$ satisfies $\|x\|_H = 1$ and $\|x\|_X \leq \sqrt{s}$. Let $\hat{x}$ be the solution to
\begin{equation}
\label{eq:minell0}
\min_{\tilde{x}} \|\tilde{x}\|_X
\quad \text{subject to} \quad A \tilde{x} = A x \quad \text{and} \quad \|\tilde{x}\|_H = 1 \pl.
\end{equation}
Since $x$ is feasible for the program in \eqref{eq:minell0}, we have $\|\hat{x}\|_X \leq \|x\|_X \leq \sqrt{s}$. Moreover, $\hat{x}$ also satisfies $\|\hat{x}\|_H = 1$.
Therefore, by Corollary~\ref{cor:weak_diff}, it follows that $\|\hat{x} - x\|_H \leq 8 \delta$. Without the unit-norm constraint, the optimization in \eqref{eq:minell0} becomes a convex program. We will pursue a guarantee for this convex program in a future work.
}
\end{rem}

\begin{rem}
{\rm
We may even replace this by
 \[ \min \|\tilde{x}\|_X \quad \mbox{\rm subject to }  \|A\tilde{x}-Ax\|< K \delta \quad \|\tilde{x}\|_H \lel 1 \pl .\]
Then we conclude that $\|\hat{x}-x\|_H\kl (K+2\sqrt{2})\delta$, and hence we can allow small random errors.
}
\end{rem}

Next we show that the MRIP holds for group structured measurements. To avoid redundancy, we illustrate a single example where the sparsity model is given with a polytope $K$.

\begin{lemma}\label{mmrip}
Let $H = \ell_2^N$ and $K$ be an absolute convex hull of $M$ points, and $K$ have enough symmetry with an isotropic affine representation $\sigma: G \to O_N$, $X$ be the Banach with unit ball $K$, $g_1,\dots,g_m$ be independent copies of a Haar-distributed random variable $g$ on $G$, $s_{\max}(K)$ be defined in \eqref{smax}, and $u:X\to\ell_2^d$ satisfy $\mathrm{tr}(u^*u) = N$. Then there exists a numerical constant $c$ such that $A = \frac{1}{\sqrt{m}} (u\si(g_j))_{1\leq j \leq m}$ satisfies the MRIP with distortion $\tilde{\delta}$ at level $\tilde{s}$ with probability $1-\zeta$ provided
\[
m \gl c \tilde{\delta}^{-2} \tilde{s} (1+\ln N) (1+\ln(\zeta^{-1})) (1+\ln m)^3 (1+\ln (s_{\max}(K))) \|u:X\to\ell_2^d\|^2 \pl .
\]
\end{lemma}

\begin{proof}
Let $l \in \nz$. Note that
\[
\frac{2^l \tilde{s}}{2^l \tilde{\delta}^2} \lel \frac{\tilde{s}}{\tilde{\delta}^2} \pl .
\]
Therefore, by \cite[Theorem~5.1]{junge2016ripI},
\[
\mathbb{P}\left(
\sup_{\|x\|_0\le 2^l \tilde{s}, ~ \|x\|_2 = 1} \Big|\frac{1}{m}\sum_{j=1}^m \|u(\si(g_j)x)\|_2^2-\|x\|_2^2\Big|
\geq \max(2^{l/2}\tilde{\delta},2^l\tilde{\delta}^2)
\right) \leq \frac{\zeta}{\lceil\log_2(N/\tilde{s})\rceil}
\]
holds provided
\[
m \gl c \tilde{\delta}^{-2} \tilde{s} (1+\ln N)
\max\Big( (1+\ln m)^3 , \ln(\zeta^{-1}) \Big) \|u:X\to\ell_2^d\|^2 \pl .
\]
Since $l$ was arbitrary, applying the union bound over all $l$ satisfying
$\lfloor -\log_2 s \rfloor \leq \l \leq \lceil \log_2 (s_{\max}(K)/s) \rceil$ gives the assertion.\qd

\section{Group structured measurements with further randomization}
\label{sec:composition}

In the Part I paper \cite{junge2016ripI}, we have shown that the number of randomly sampled group structured measurements $m$ for the RIP scales near optimally for certain sparsity models (e.g., sparsity models with respect to Banach spaces $X = \ell_1^N$ or $X = S_1^n$). However, in general, one needs a larger number of measurements for the group structured case than that for Gaussian measurements.
In this section, we propose two different ways to improve the sample complexity for the RIP of randomly sampled group structured measurements.
To this end, we first recall the optimal result on the number of Gaussian measurements for the RIP. Gordon's lemma \cite{gordon1988milman} shows that a minimal number of Gaussian measurements provide the dimensionality reduction of an arbitrary set $D$ consisting of unit-norm vectors.

\begin{lemma}[Gordon's escape through the mesh \cite{gordon1988milman}]\label{fgau}
Let $0 < \delta < 1$, $D$ be a subset of the unit sphere $\mathbb{S}^{N-1}$, and $\bm{\xi}_1,\dots,\bm{\xi}_m$ be independent copies of a standard Gaussian random vector $\bm{\xi} \sim \mathcal{N}(0,I_N)$.
Then
\[
\mathbb{P}\left(
\sup_{x \in D} \Big| \frac{1}{m} \sum_{j=1}^m |\langle \bm{\xi}_j,x\rangle|^2 - \|x\|_2^2 \Big|
\geq \delta
\right) \leq \zeta
\]
holds provided
\[
m \geq \delta^{-2} \left(\ell(D) + \sqrt{2 \ln(2/\zeta)}\right)^2 \pl,
\]
where $\ell(D)$ denotes the Gaussian width of $D$ defined by
\begin{equation}
\label{gauwidth}
\ell(D) := \ez \sup_{x \in D} \langle \bm{\xi}, x \rangle \pl.
\end{equation}
\end{lemma}

\begin{rem}{\rm
The Gaussian width in \eqref{gauwidth} satisfies $\ell(D) = \ell(\absc D)$, where $\absc D$ denotes the absolute convex hull of $D$. It also coincides with the Gaussian-summing norm \cite{diestel1995absolutely} (also known as the $\ell$-norm) of the identity operator from $\ell_2^N$ to $Y$, where $Y$ is the Banach space with unit ball $\absc D$. Our notation for the Gaussian width is motivated by this interpretation.
}\end{rem}

In the remainder of this section, we consider $D = K_s = \sqrt{s} K \cap \mathbb{S}^{N-1}$ for a fixed convex body $K\subset B_2^N$. Then it follows that $\ell(D) \leq \ell(\sqrt{s} K) = \sqrt{s} \ell(K)$.

\subsection{Group structured measurements followed by a small subgaussian matrix}

Suppose that $K$ has enough symmetries with an isotropic, affine representation $\si$ on a compact group $G$. Let $X$ be the Banach with unit ball $K$. A fixed map $u:X\to\ell_2^d$ can be viewed as an instrument to probe data from $X$.

We consider a situation where a sufficient number of group structured measurements are stored in a data bank but the original data is no longer accessible. We can further reduce the amount of data by applying a small Gaussian matrix to the vectors with the obtained measurements. The next result shows that the measurements obtained by the two-step approach by the composition operator provide dimensionality reduction, which is as good as that by pure Gaussian measurements in Gordon's lemma.

\begin{theorem}
\label{composition}
Let $1\le p\le2$. Let $K$, $X$, $G$, $\si$, and $u$ be as above.
Let $v_j = u\si(g_j)$ for $j=1,\dots,M$, where $g_1,\dots,g_M$ are independent copies of a Haar-distributed random variable in $G$, and $\bm{\xi}_1,\dots,\bm{\xi}_m$ be independent copies of a standard Gaussian random vector $\bm{\xi} \sim \mathcal{N}(0,I_M)$.
Let $\al_d$ be an 1-homogeneous and $G$-invariant function on $B(X,\ell_2^d)$.
Then there exists a numerical constant $c$ such that $A = \frac{1}{\sqrt{M}} (v_j)_{1\le j\le M}$ satisfies
\[
\mathbb{P}\left(
\sup_{x\in K_s} \Big|\frac{1}{m}\sum_{l=1}^m |\langle\bm{\xi}_l,Ax\rangle|^2-\|x\|_2^2\Big|
\geq \max(\delta,\delta^2)
\right) \leq \zeta
\]
provided that
\begin{align}
\label{composition:cond1} m & \geq 3 \delta^{-2} \Big(\sqrt{s} \ell(K) + \sqrt{2\ln(2/\zeta)}\Big)^2 \\
\label{composition:cond2} \frac{M}{1+\ln(NMd)} & \gl c\|u\|_{S_\infty}^2 \\
\label{composition:cond3} \frac{M^{1/p}}{(1+\ln M)^{e(p)/2}} & \gl c M_{p,\al_d}(K) \sqrt{s} \delta^{-1} \al_d(u) \\
\label{composition:cond4} M & \gl c \delta^{-2} s \ln(\zeta^{-1}) \|u:X\to\ell_2^d\|^2 \pl ,
\end{align}
where $e(\cdot)$ is defined by $e(p) = 1$ for $1 \leq p < 2$ and $e(p) = 3$ for $p=2$.

In particular for a large enough $M$ a low dimensional image of $K$ with ``Gordon optimal number of measurements'' can be reproduced by group translations of a given instrument $u$.
\end{theorem}

In general, the number of group translations, $M$, determined by Theorem~\ref{composition} is larger than the number of measurements, $m$, at the final stage. However, $M$ is still significantly smaller than the ambient dimension $N$ and the size of the Gaussian matrix, $m$-by-$M$, is small compared to the size $m$-by-$N$, of a pure Gaussian measurement matrix. Since group structured measurement operators often have fast implementation (e.g., partial Fourier \cite{fessler2003nonuniform} and partial Gabor \cite{sondergaard2012efficient}), the composition measurement system is scalable and can be useful for high dimensional data in particular with a sparse $u$.

The proof of Theorem~\ref{composition} consists of a sequence of supporting lemmas.
For the first lemma, we need a standard application of Rosenthal's inequality.

\begin{lemma}\label{Ros} Let $\si:G\to O_N$ be an affine isotropic representation, where $O_N$ denotes the orthogonal group, $g_1,\dots,g_M$ be independent copies of a Haar-distributed random variables $g$ in $G$, and $u:\cz^N\to \cz^d$ be a linear map such that $\mathrm{tr}(u^*u)=N$.
Then
\[
\ez \Big\| \frac{1}{M}\sum_{j=1}^M \si(g_j)^*u^*u\si(g_j) - \mathrm{Id} \Big\|_{S_\infty}
\kl \frac{C \|u\|_{S_\infty}\sqrt{1+\ln (NMd)}}{\sqrt{M}}
\Big(1+\frac{\|u\|_{S_\infty}\sqrt{1+\ln (NMd)} }{\sqrt{M}}\Big) \pl
\]
for a numerical constant $C$.
 \end{lemma}

\begin{proof}
Since the action is isotropic, we see that  $\ez \si(g)u^*u\si(g)^*=\mathrm{Id}$ and
\[
\ez \si(g)u^*u\si(g)\si(g)^*u^*u\si(g)
\lel \ez \si(g)u^*uu^*u\si(g)
\lel \frac{\mathrm{tr}(u^*uu^*u)}{N} \mathrm{Id} \pl.
\]
With the help of the noncommutative Rosenthal inequality (see \cite[Theorem~0.4]{junge2013noncommutative}) this implies
\begin{align*}
&\ez \Big\|\sum_{j=1}^M \si(g_j)^*u^*u\si(g_j) - \ez \si(g_j)^*u^*u\si(g_j) \Big\|_{S_p} \\
&\le C p M^{1/p} \|u^*u\|_{S_p}
+ C \sqrt{pM} \|\ez gu^*uu^*ug\|_{S_{p/2}}^{1/2} \\
&\le C p (Md)^{1/p}\|u\|_{S_\infty}^2
+ C (pM/N)^{1/2} N^{1/p} \|u\|_{S_4}^{2} \\
&\le C p (Md)^{1/p}\|u\|_{S_\infty}^2
+ C (pM/N)^{1/2} N^{1/p} \|u\|_{S_\infty}\|u\|_{S_2}    \\
&\le C \|u\|_{S_\infty}(pM)^{1/2} N^{1/p}
(\sqrt{p}M^{-1/2}(Md/N)^{1/p}\|u\|_{S_\infty} +1) \pl .
\end{align*}
In case $Md/N\le 1$, we choose $p=\ln N$ and find
\[
\ez \Big\|\frac{1}{M}\sum_{j=1}^M \si(g_j)^*u^*u\si(g_j)-I_N\Big\|_{S_\infty}
\kl \frac{C \|u\|_{S_\infty}\sqrt{1+\ln N}}{\sqrt{M}}
\Big(1+\frac{\|u\|_{S_\infty}\sqrt{1+\ln N}}{\sqrt{M}}\Big) \pl .
\]
For $Md/N\gl 1$ we simply choose $p=1+\ln(Md/N)+\ln N$  and get
\[
\ez \Big\|\frac{1}{M}\sum_{j=1}^M \si(g_j)^*u^*u\si(g_j)-I_N\Big\|_{S_\infty}
\kl \frac{C \|u\|_{S_\infty}\sqrt{1+\ln (Md)}}{\sqrt{M}}
\Big(1+\frac{\|u\|_{S_\infty}\sqrt{1+\ln (Md)}}{\sqrt{M}}\Big) \pl.
\]
Then the assertion follows.  \qd

\begin{cor}\label{invert}
Let $u$, $K$, $A$ be as in Theorem~\ref{composition}.
Suppose that \eqref{composition:cond2} holds.
Then $A$ is invertible on its range and in particular satisfies
\[
\ell(A(K)) \underset{\sqrt{3}}{\sim} \ell(K) \pl .
\]
\end{cor}

\begin{proof} By Lemma \ref{Ros}, there exists $c > 0$ such that the assumption implies
\[
\|A^*A-\mathrm{Id}\|_{S_\infty}<\frac{1}{2} \pl .
\]
Then $A$ is injective.
Furthermore it follows that $\|A\|_{S_\infty} \leq \sqrt{3/2}$ and $\|A^\dagger\|_{S_\infty} \leq \sqrt{2}$, where $A^\dagger$ denotes the Penrose-Moore pseudo inverse of $A$.
Therefore $\ell(A(K))\sim_{\sqrt{3}}\ell(K)$.\qd

Now by combining the above lemmas, we prove Theorem~\ref{composition}.
\begin{proof}[Proof of Theorem~\ref{composition}]
First, by \cite[Theorem~5.1]{junge2016ripI}, there exists a numerical constant $c$ such that $A$ satisfies the RIP on $K_s$ with constant $\delta$ with probability $1-\zeta/2$ provided that \eqref{composition:cond3} and \eqref{composition:cond4} hold.
Next, due to Lemma \ref{invert}, we may assume that $\|A\|_{S_\infty} \le \sqrt{3/2}$.
Therefore Lemma~\ref{fgau} applies to $A(K)$ and we find that \eqref{composition:cond1} implies
\[
\mathbb{P}\left(
\sup_{x\in K_s}
\left|\frac{1}{m}\sum_{l=1}^m |\langle\xi_l,Ax\rangle|^2-\|Ax\|_2^2\right|
\geq \max(\delta/2,\delta^2/4)
\right) \leq \zeta \pl.
\]
Then the triangle inequality implies the assertion.
\qd

\subsection{Group structured measurement with random sign} $\atop$

Here we demonstrate that one can reduce the number of group structured measurements for the RIP by certain choices of groups so that it is comparable to the (widely accepted optimal) result by Gaussian measurements. More precisely, adopting the multiresolution RIP result by Oymak et al. \cite{oymak2015isometric}, we show that the combination of the group structured measurement operator and the diagonal operator with random sign achieves the ``Gordon optimal number of measurements''. We start with the observation that the group structured measurement generalizes the partial Fourier measurement and satisfies the multiresolution RIP.
Here, we use the original definition of the multiresolution RIP \cite{oymak2015isometric}, which is a special case of Definition~\ref{mrip_gen}.

\begin{defi}[{Multiresolution RIP \cite{oymak2015isometric}}]
We say that $A:\ell_2^N\to\ell_2^m$ satisfies the Multiresolution Restricted Isometry Property (MRIP) with distortion $\delta > 0$ at sparsity level $s \geq 1$ if
\[
\sup_{\|x\|_0 \leq 2^l s} |\|Ax\|_2^2 - \|x\|_2^2| \leq \max(2^{l/2} \delta,~2^l \delta^2)
\]
for all $l=0,1,\dots,\lceil \log_2 (N/s) \rceil$.
\end{defi}

We fix a group $G$ with an isotropic affine representation $\si:G\to O_N$ such that $\si(g)$ is an isometry on $\ell_1^N$ and $\ell_2^N$ simultaneously for all $g \in G$.

\begin{theorem}
\label{Oy1sparse}
Let $G$ be a group action with an affine representation $\si:G\to O_N$ such that $\si(g)$ is an isometry on both $\ell_1^N$ and $\ell_2^N$ for all $g \in G$. Let $\eta\in\ell_2^N$ be a vector of $\|\eta\|_2=\sqrt{N}$, $g_1,\dots,g_m$ be independent copies of a Haar-distributed random variable $g$ in $G$, and $v_j:\ell_1^N\to\cz$ be given by $v_jx = \langle\si(g_j)^*\eta,x\rangle$ for $j=1,\dots,m$.
Suppose in addition that $D_\eps \in \mathbb{R}^{N \times N}$ is a diagonal matrix whose diagonal entries are from a Rademacher sequence $(\eps_j)_{1\le j\le N}$.
Let $\T \subset \mathbb{S}^{N-1}$.
Then there exists a numerical constant $c$ such that
\begin{align} \label{ggg}
\mathbb{P}\left(
\sup_{x\in \T} \Big|\frac{1}{m}\sum_{j=1}^m |\langle D_{\eps}\si(g_j)^*\eta,x\rangle|^2 - \|x\|_2^2\Big| \geq \max(\delta,\delta^2)
\right) \leq \zeta
\end{align}
provided
\[
m \gl c \delta^{-2} \ell(\T)^2 (1+\ln(\zeta^{-1})) (1+\ln N) \max\Big((1+\ln m)^3, (1+\ln(\zeta^{-1}))\Big) \|\eta\|_{\infty}^2 \pl .
\]
\end{theorem}

\begin{proof}[Proof of Theorem~\ref{Oy1sparse}]
The assertion follows by combining Lemma~\ref{mmrip} for $K=B_1^N$
and \cite[Theorem~3.1]{oymak2015isometric}
for $\tilde{s} = C_1(1+\ln(\zeta^{-1}))$ and $\tilde{\delta} = \frac{\delta}{C_2 \ell(\mathcal{T})}$ with some numerical constants $C_1$ and $C_2$. \qd

\begin{cor}\label{Oy}
Under the hypothesis of Theorem~\ref{Oy1sparse}, suppose in addition that $K \subset B_2^N$ is convex. Let $K_s = \sqrt{s} K \cap \mathbb{S}^{N-1}$.
Then there exists a numerical constant $c$ such that
\[
\mathbb{P}\left(
\sup_{x\in K_s} \Big|\frac{1}{m}\sum_{j=1}^m |\langle D_{\eps}\si(g_j)^*\eta,x\rangle|^2 - \|x\|_2^2\Big| \geq \max(\delta,\delta^2)
\right) \leq \zeta
\]
holds provided
\begin{equation}
\label{eq:Oy:cond}
m \gl c \delta^{-2} s \ell(K)^2 (1+\ln(\zeta^{-1})) (1+\ln N) \max\Big((1+\ln m)^3, (1+\ln(\zeta^{-1}))\Big) \|\eta\|_{\infty}^2 \pl .
\end{equation}
\end{cor}

\begin{proof} This follows from $\ell(K_s)\le \sqrt{s}\ell(K)$. \qd

\begin{rem}
{\rm
One may improve Corollary~\ref{Oy} by considering a better estimate on $\ell(K_s)$ given by
\[
\ell(K_s)
\lel \ez \sup_{x\in K_s} |\langle\bm{\xi},x\rangle|
\kl \inf_{\mathrm{Id}=a+b} \|a\|_{S_2}+\ell(b(K_s)) \pl ,
\]
which is smaller than $\sqrt{s} \ell(K)$.
}
\end{rem}

Indeed, for some unconditional Banach sequence space norms, the diagonal operator with random sign can be absorbed into the group action, which is stated in the following corollary.

\begin{cor} \label{largecross}
Consider an affine representation $\tilde{\si}$ of $\widetilde{G}=\{-1,1\}^N\rtimes \zz_N$ via diagonal matrices and shift matrices. Suppose that $\tilde{g}_1,\dots,\tilde{g}_m$ are independent copies of a random variable $\tilde{g}$ in $\widetilde{G}$ with respect to the Haar measure and $K \subset B_2^N$ is $\widetilde{G}$-invariant.
Let $K_s = \sqrt{s} K \cap \mathbb{S}^{N-1}$.
Then there exists a numerical constant $c$ such that
\[
\mathbb{P}\left(
\sup_{x\in K_s} \Big|\frac{1}{m}\sum_{j=1}^m |\langle \tilde{\si}(\tilde{g}_j)^*\eta,x\rangle|^2 - \|x\|_2^2\Big| \geq \max(\delta,\delta^2)
\right) \leq \zeta
\]
provided that the condition in \eqref{eq:Oy:cond} holds.
\end{cor}

\begin{proof}
In \cite[Section~4]{junge2016ripI}, we have shown that $\tilde{\si}$ is an isotropic affine representation of $\widetilde{G}$.
Then we note that thanks to the group invariance for all $r \in \mathbb{N}$
\begin{align*}
& \Big(\ez_{\widetilde{G}^m} \sup_{x\in K_s} \Big|\frac{1}{m}\sum_{j=1}^m |\langle \tilde{\si}(\tilde{g}_j)\eta,x\rangle|^2-\|x\|^2\Big|^r\Big)^{1/r} \\
&= \Big(\ez_{\eps} \ez_{G^m}\sup_{x\in K_s} \Big|\frac{1}{m}\sum_{j=1}^m |\langle D_{\eps}\si(g_j)\eta,x\rangle|^2-\|x\|^2\Big|^r\Big)^{1/r} \\
&= \Big(\ez_{G^m} \ez_{\eps}
\sup_{x\in K_s} \Big|\frac{1}{m}\sum_{j=1}^m |\langle \si(g_j)\eta,D_{\eps}x\rangle|^2-\|x\|^2\Big|^r\Big)^{1/r} \pl .
\end{align*}
Now it suffices to apply Corollary \ref{Oy}. \qd

\begin{rem}
{\rm
Note that one can use use different kinds of randomness and have different degree of generality for group structured measurements. In i) we can work with an arbitrary group $G$ and are not using good concentration inequalities. In ii) we have to restrict to subgroups of isometries of $\ell_1^N$, and that can be very restrictive. It is also interesting to consider the sampling sets in i) and ii). In i) we sample from the orbit
\[ \si(G)^*\eta\]
of cardinality at most $|G|$, and ii) from
\[ \{-1,1\}^N\si(G)^*\eta  \]
of cardinality at most $2^N|G|$. With respect to this criterion Fourier sampling only uses $N$ vectors to choose from for $G=\zz_N^2$ with the affine representation. The particular $\eta=[1,\dots,1]^\top \in \rz^N$ also produces only an orbit of $N$ vectors. We suspect that our methods can be modified to other type of randomness where a random set is chosen out of the orbit. Although the gaussian case provides the ``best'' estimate it requires considerably more ``randomness''.
From this perspective, the improvement given by Corollary~\ref{largecross} for a nice group $\widetilde{G}=\{-1,1\}^N\rtimes \zz_N$ over the choice of $\zz_N^2$ comes at a penalty that the cardinality of the orbit is larger ($|\widetilde{G}^m|=2^{Nm}N^m$ versus $|G^m|=N^{2m}$). This may increase the cost of implementation.
}
\end{rem}

Finally, we conclude this section with an application of Corollary~\ref{Oy} to low-rank tensors.
Let $K_s$ denote the set of rank-$s$ tensors in $(\ell_2^n)^{\otimes_\pi^d}$.
In the Part I paper \cite[Theorem~7.3]{junge2016ripI}, we showed that the group structured measurement with a standard Gaussian instrument $\bm{\xi}$ satisfies
\[
\mathbb{P}\left( \sup_{x\in K_s} \Big|\frac{1}{m}\sum_{j=1}^m |\langle \si(g_j)^*\bm{\xi},x\rangle|^2 - \|x\|_2^2\Big| \geq \max(\delta,\delta^2) \right) \leq \zeta
\]
provided
\[
m \gl c \delta^{-2} s (1+\ln m)^3 (1+3nd(1+\ln d)+\ln(\zeta^{-1}))^2 \pl .
\]
We compare this estimate to that by the group structured measurement operator after applying random sign. Let $(\eps_j)_{1\le j\le N}$ be a Rademacher sequence and $D_\eps$ be the corresponding diagonal operator with random sign. Then applying the fact that a standard Gaussian $\bm{\xi} \in \rz^N$ satisfies
\[
\mathbb{P}\left(\|\bm{\xi}\|_\infty \geq \sqrt{2(1+\ln N+\ln(\zeta^{-1}))}\right) \leq \zeta \pl
\]
to Corollary~\ref{Oy}, it follows that
\[
\mathbb{P}\left(
\sup_{x\in K_s} \Big|\frac{1}{m}\sum_{j=1}^m |\langle D_\eps \si(g_j)^*\bm{\xi},x\rangle|^2 - \|x\|_2^2\Big| \geq \max(\delta,\delta^2)
\right) \leq \zeta
\]
provided
\[
m \gl c \delta^{-2} s nd(1+\ln d) (1+\ln(\zeta^{-1})) (1+d\ln n) ((1+\ln m)^3+\ln(\zeta^{-1})) (1+d\ln n+\ln(\zeta^{-1})) \pl .
\]

To simplify the expressions for the number of measurements, let us choose $\zeta$ not too small so that $\ln(\zeta^{-1})$ is dominated by the other logarithmic terms and then ignore the logarithmic terms. In Table~\ref{tab:comparison}, we compare different measurement operators in terms of the simplified number of measurements up to a logarithmic factor that provide the RIP on $K_s$ with high probability.

\begin{table}[h]
  \caption{Number of measurements for the RIP on rank-$s$ tensors in $(\ell_2^n)^{\ten_d^\pi}$.}
  \label{tab:comparison}
  \begin{tabular}{l|c|c}
  Measurement type & $m$ up to a log factor & size of orbit by group actions  \\\hline\hline
  Pure Gaussian measurements & $snd$ & \\\hline
  Group structured measurements & \multirow{2}{*}{$sn^2d^2$} & \multirow{2}{*}{$n^{2d}$} \\
  by a random generator & & \\\hline
  Group structured measurements & \multirow{3}{*}{$snd^3$} & \multirow{3}{*}{$2^{n^d} n^d$} \\
  by a random generator & & \\
  with random sign & &
  \end{tabular}
\end{table}

Then the number of group structured measurements for RIP is roughly $snd^3$ whereas the one without random sign is $sn^2d^2$. Usually, $d$ is smaller than $n$ and the random sign reduces the number of measurements in this case. For the group structured measurement operator after random sign, $m$ can be further reduced by factor $d$ if a deterministic vector (e.g., $\eta=[1,\dots,1]^\top \in \rz^N$) is used instead. The Gaussian measurement case has optimal scaling of $snd$. But its implementation may be impractical for high dimensional data. Recall that in this comparison the group structured measurements also used random instrument. However, the requirement on this random instrument is rather mild and does not require strong concentration property of the Gaussian matrix.

On the other hand, note here that the transformations $\si(g)$ preserve both the convex body $K$ and the $\ell_2$-norm. However, although the diagonal operator $D_{\eps}$ also preserves the $\ell_2$-norm, unlike $\si(g)$, $D_{\eps}$ does not preserve $K$ in general. Also note that there are applications where random sign cannot be implemented.

\section{RIP in infinite dimension}
\label{sec:infdim}

In this section, we present the RIP on infinite dimensional sparsity models. A general version is described below in the Fourier analysis framework, followed by illustrations on two specific examples of the model.

\subsection{RIP on infinite dimensional sparsity models}$\atop$

We will consider various (semi-) norms on $L_2(0,1)$ defined by using the Fourier series representation. Let $(\hat{f}(k))_{k\in\zz}$ denote the Fourier series of the periodization of $f \in L_2(0,1)$, i.e.
\[
\hat{f}(k) = \langle \psi_k, f \rangle = \int_0^1 \overline{\psi_k(t)} f(t) dt, \quad k \in \zz \pl ,
\]
where $\psi_k$s are complex sinusoid functions defined by
\begin{equation}
\label{defpsik}
\psi_k(t)=e^{2\pi \mathfrak{i} k t}, \quad t \in [0,1) \pl.
\end{equation}

Let $w = (w_k)_{k\in\zz}$ be a nonnegative sequence and define a weighted (semi)norm
\begin{equation}
\label{def2wnorm}
\|f\|_{2,w}^2 \lel \sum_{k \in \zz} w_k |\hat{f}(k)|^2 \pl .
\end{equation}
This weighted (semi)norm is reminiscent of usual Besov norms. For example, if $w_k=k$ for $k \in \zz$, then it follows that $\|f\|_{2,w}=\|f'\|_{L_2}$ for all $f \in C^1$. Let $H_w$ denote the (semi)normed space equipped with $\|\cdot\|_{2,w}$. If $w_k > 0$ for all $k\in\zz$, then $\|\cdot\|_{2,w}$ is a valid norm and $H_w$ is a Hilbert space. But we are also interested in the case where $w$ is finitely supported and $\|\cdot\|_{2,w}$ is just a seminorm.

In the Part I paper \cite{junge2016ripI}, we defined that a vector $x \in H$ is $(K,s)$-sparse if $\|x\|_X \leq \sqrt{s} \|x\|_H$ for a Hilbert space $H$ and a Banach space $X$ induced from a convex set $K \subset H$ such that the unit ball in $X$ is $K$. One may note that the main results there were derived without using the fact that the Hilbert space norm is definite, i.e. $\|x\|_H = 0$ implies $x=0$, and remain valid when the Hilbert space $H$ is replaced by a seminormed space $H_w$. In particular, \cite[Theorem~2.1]{junge2016ripI} applies to an infinite dimensional sparsity model in $L_2(0,1)$ defined by a Banach space $L_q(0,1)$ and a (semi)normed space $H_w$ as follows.

\begin{defi}
\label{definfsl}
Let $1<q\le 2$ and fix a weight sequence $w = (w_k)_{k \in \zz}$.
We say that $f\in L_2(0,1)$ is $s$-sparse with respect to $(H_w,L_q(0,1))$ if
\[
\|f\|_{L_q} \kl \sqrt{s} \|f\|_{2,w} \pl .
\]
\end{defi}

As in the Part I paper \cite{junge2016ripI}, we consider random group structured measurements. Let $\tau_t$ be a group action on $L_2(0,1)$ that maps $f$ to its translation to the right by $t \in [0,1)$ modulo $1$ and $u:L_2(0,1)\to \ell_2^d$ be a linear operator that takes a Hilbert space valued measurement. Then group structured measurements of $f$ are generated as $(u(\tau_{t_j}f))_{j=1}^m$ where $t_1,\dots,t_m$ are independent copies of a uniform random variable on $[0,1)$.

We are interested in a specific $H_w$ with the weight sequence determined by $u$ as
\begin{equation}
\label{defwk}
w_k \lel \|u(\psi_k)\|_2^2, \quad k \in \zz \pl .
\end{equation}
Then it follows that
\begin{equation}
\label{individual_expectation}
\ez \|u(\tau_{t_j}f)\|_2^2 = \|f\|_{2,w}, \quad \forall j=1,\dots,m \pl .
\end{equation}
Indeed, since
\begin{align*}
\ez \langle\psi_k, (\tau_t u^*u \tau_t) \psi_j \rangle
&= \int_0^1 e^{-2\pi \mathfrak{i}(k-j)t} \langle\psi_k,(u^*u)(\psi_j)\rangle dt
\lel \delta_{kj} \langle\psi_k, (u^*u)(\psi_j)\rangle \pl ,
\end{align*}
where $\delta_{kj}$ denotes the Kronecker delta, it follows that $\ez \tau_t^* u^*u\tau_t$ is a Fourier multiplier such that
\[
(\ez \tau_t^* u^*u\tau_t)(f) = \sum_{k\in\zz} w_k \psi_k \langle\psi_k,f\rangle, \quad
f\in L_2(0,1) \pl .
\]

In this setup, the main theorem in the Part I paper \cite[Theorem~2.1]{junge2016ripI} provides the following corollary.
\begin{cor}
\label{infshift}
Let $1<q\le 2$, $0<\zeta<1$, and $\delta>0$.
Let $\|\cdot\|_{2,w}$ be defined by \eqref{def2wnorm} where $(w_j)_{j\in\zz}$ is given from $u$ as \eqref{defwk}.
Suppose that $t_1,\dots,t_m$ are independent copies of a uniform random variable on $[0,1)$. Then there exists a numerical constant $c$ such that
\begin{equation}
\label{infshift:rip}
\mathbb{P}\left(
\sup_{\begin{subarray}{c} \|f\|_{L_q}\le\sqrt{s} \\ \|f\|_{2,w}=1 \end{subarray}}
\Big|\frac{1}{m}\sum_{j=1}^m \|u(\tau_{t_j}f)\|_2^2 -\|f\|_{2,w}^2\Big|
\geq \max(\delta,\delta^2)
\right) \leq \zeta
\end{equation}
provided
\[
m \gl c \delta^{-2} s \max\Big((q')^3 (1+\ln d)^3 (1+\ln m)^3, \ln(\zeta^{-1})\Big)
\Big\|\Big(\sum_{l=1}^d |u^*(\bm{e}_l)|^2\Big)^{1/2}\Big\|_{L_{q'}}^2 \pl .
\]
Here, $\bm{e}_1,\dots,\bm{e}_d$ denote the standard basis vectors in $\rz^d$.
\end{cor}

\begin{proof}
The proof is similar to that of \cite[Theorem~5.3]{junge2016ripI} and consists of verifying the conditions of \cite[Theorem~2.1]{junge2016ripI}. We apply \cite[Theorem~2.1]{junge2016ripI} to $H_w$ (instead of $H$) with $p=2$ and the 1-homogeneous function $\al_d(\cdot)$ given by
\[
\al_d(u) = \Big\|\Big(\sum_{l=1}^d |u^*(\bm{e}_l)|^2\Big)^{1/2}\Big\|_{L_{q'}} \pl.
\]

First, instead of the isotropy, from \eqref{individual_expectation} it follows that
\[
\ez \frac{1}{m} \sum_{j=1}^m \|u(\tau_{t_j} f)\|_2^2 = \|f\|_{2,w} \pl .
\]
Second, $L_q(0,1)$ is a lattice and the complexity of its unit ball $K$ is upper bounded by \cite[Theorem~3.8]{junge2016ripI}, where the type 2 constant of $L_{q'}(0,1)$ is no larger than $\sqrt{q'}$ \cite[Lemma~3]{carl1985inequalities}.
Lastly we note
\begin{align*}
\|u\|
&= \sup_{\|f\|_{L_q} \leq 1} \Big(\sum_{l=1}^d |\langle \bm{e}_l, u(f)\rangle|^2\Big)^{1/2}
= \sup_{\|f\|_{L_q} \leq 1} \Big(\sum_{l=1}^d |\langle u^*(\bm{e}_l), f\rangle|^2\Big)^{1/2} \\
& \leq \sup_{\|f\|_{L_q} \leq 1} \Big\langle \Big(\sum_{l=1}^d |u^*(\bm{e}_l)|^2\Big)^{1/2}, f \Big\rangle
= \al_d(u) \pl .
\end{align*}
Then the assertion follows from \cite[Theorem~2.1]{junge2016ripI}.
\qd

If $w_k = 1$ for all $k \in \zz$, the weighted norm reduces to the usual $L_2$-norm and $H_w$ becomes the Hilbert space $L_2(0,1)$. However, since \eqref{defwk} is not satisfied for this weight sequence with $d < \infty$, Corollary~\ref{infshift} does not apply to this case. However, there exist interesting instances of $H_w$ where Corollary~\ref{infshift} provides efficient compression of data in infinite dimension. In the remainder of this section, we will illustrate Corollary~\ref{infshift} for two specific $H_w$s with motivating applications.

\subsection{Preserving truncated seminorms}$\atop$

In the first example of $H_w$, we consider $\|\cdot\|_{w,2}$ given as a truncated seminorm with the weight sequence $w = (w_k)_{k\in\zz}$ given by
\begin{equation}
\label{finsuppwk}
w_k =
\begin{cases}
1 & -N \leq k < N \pl , \\
0 & \mathrm{otherwise} \pl .
\end{cases}
\end{equation}
In this setup, Corollary~\ref{infshift} implies that group structured measurements preserve the $\ell_2$-norm of a subsequence of the Fourier series of $f$ restricted to $[-N,N) \cap \zz$. This result combined with the arguments in Section~\ref{sketch} shows that a subsequence of the Fourier series within the given interval can be reconstructed from the described group structured measurements.

We will apply Corollary~\ref{infshift} to specific measurement maps $u:L_q(0,1)\to\ell_2^d$ such that the weight sequence $(w_k)_{k\in\zz}$ from $u$ by \eqref{defwk} satisfies \eqref{finsuppwk}. Before discussing the number of measurements for the RIP, we elaborate on the sparsity model to get a meaningful physical interpretation of the sparsity level.

\subsubsection{Smooth and sparse signals}

In the canonical sparsity model in $\rz^n$, where the sparsity level counts the number of nonzero entries, the ratio of the sparsity level $s$ of an instance to the dimension of the ambient space $n$ takes a value in the unit interval $[0,1]$ and clearly indicates how sparse the instance is. However, the sparsity level $s$ of $f \in L_2(0,1)$ in Definition~\ref{definfsl} with the weight sequence in \eqref{finsuppwk} does not provide such an interpretation. Here we will derive a subset of the sparsity model involving two interpretable parameters. First let us indicate that the new notion of sparsity can be viewed as combination of two sparsity conditions.

\begin{lemma}\label{ooo} Let $1<q\le 2$. Suppose that $f\in L_2(0,1)$ satisfies
\begin{enumerate}
  \item[i)] $\|f\|_{L_2} \kl \al \|f\|_{2,w}$;
  \item[ii)] $f$ is supported on a set of measure $\gamma$;
\end{enumerate}
Then $f$ is $s$-sparse with respect to $(H_w,L_q(0,1))$ with sparsity level $s = \al^2 \gamma^{2/q-1}$.
\end{lemma}

\begin{proof} Let $E$ be the support of $f$ and $1_E(t)$ be the indicator function of $E$. Then we have
\[
\|f\|_{L_q}^q \kl \int 1_E(t) |f(t)|^q dt
\kl \la(E)^{1-q/2} \Big(\int |f(t)|^{q(2/q)}dt\Big)^{q/2} \pl .
\]
This implies
\[ \|f\|_{L_q}\kl \gamma^{1/q-1/2}\|f\|_{L_2}\kl \gamma^{1/q-1/2}\al \|f\|_{2,w} \pl .\]
Thus by definition the sparsity level $s$ is at most $\gamma^{2/q-1}\al^2$. \qd

When $\|\cdot\|_{w,2}$ is given by the weight sequence in \eqref{finsuppwk}, the first condition in Lemma~\ref{ooo} implies that the Fourier coefficients within $[-N,N) \cap \zz$ consume a fraction of the $L_2$-norm of $f$. Due to the Heisenberg uncertainty principle, $f \in L_2(0,1)$ cannot have small support in both the time and Fourier domains simultaneously. However, a certain notion of sparsity can imply the concentration of a fraction of the $L_2$ norm within a small support in the Fourier domain. This is shown in the following lemma.

\begin{lemma}\label{smooth}
Let $1<q\leq 2$, $N \in \nz$, $(w_k)_{k\in\zz}$ be defined in \eqref{finsuppwk}, and $H_w$ be defined as above.
Define
\begin{equation}\label{KCgamma}
K_{\rho,\gamma}\lel \{f \in L_2(0,1) \cap C^1 \pl|\pl \|f'\|\kl \rho \|f\|_{L_2},\pl \la({\rm supp}(f))\kl \gamma \} \pl ,
\end{equation}
where $\lambda$ denotes the normalized Lebesgue measure on $(0,1)$.
Suppose that $\rho \le N/2$.
Then $f \in K_{\rho,\gamma}$ implies
\[
\|f\|_{L_q} \leq \sqrt{s} \|f\|_{2,w}
\]
for $s=(1+4\rho^2/N^2) \gamma^{2/q-1}$.
\end{lemma}

\begin{proof}
Note that
\[
\sum_{|k|\geq N} |\hat{f}(k)|^2
\kl N^{-2}\sum_k |k\hat{f}(k)|^2 \kl N^{-2}\|f'\|_{L_2} \pl .
\]
Without loss of generality, we may assume $\|f\|_{L_2}=1$, and deduce that
\[
1 \lel \|f\|_{L_2}^2 \kl \|f\|_{2,w}^2+ \frac{\rho^2}{N^2}
\]
and hence $(1-\rho^2/N^2)\kl \|f\|_{2,w}^2$.
Since $(1-t)^{-2}\le (1+4t)$ for $0<t\le1/4$, by the assumption $N \geq 2\rho$, it follows that
\[
\|f\|_{L_2}
\kl (1-\rho^2/N^2)^{-1} \|f\|_{2,w}
\kl (1+4\rho^2/N^2)^{1/2} \|f\|_{2,w} \pl .
\]
Let $\al = (1+4\rho^2/N^2)^{1/2}$.
Furthermore, since $f$ is supported on a set of measure $\gamma$, the assertion follows from Lemma~\ref{ooo}.\qd

The next lemma provides concrete examples that belong to $K_{\rho,\gamma}$.

\begin{lemma}
\label{superposition_model}
Let $\phi$ be a positive smooth and differentiable function with $\phi(0)=1$ and support contained in $[-1/2,1/2]$. Let $\phi_T(x)=T\phi(Tx)$ and $t_1,\dots,t_l$ with distance strictly bigger than $1/T$. Then
\[
f_T(t) \lel \sum_{j=1}^l \al_j \phi_T(t-t_j)
\]
satisfies
\begin{enumerate}
  \item[i)] $\la({\rm supp}(f_T))\kl \frac{l}{T}$,
  \item[ii)] $\|f_T\|_{L_p} \lel \|\phi\|_{L_p} T^{1/p'} (\sum_{j=1}^l |\al_j|^p)^{1/p}$,
  \item[iii)] $\|f_T'\|_{L_2} \lel \frac{\|\phi'\|_{L_2}}{\|\phi\|_{L_2}} T \|f_T\|_{L_2}$
\end{enumerate}
\end{lemma}

\begin{proof} We just note that $f_T$ is a sum of disjointly supported functions. This proves i) and moreover
\begin{align*}
\|f_T\|_{L_p}^p &= \sum_{j=1}^n |\al_j|^p \|\phi_T\|_{L_p}^p = \sum_{j=1}^n |\al_j|^p T^{p-1} \|\phi\|_{L_p}^p \pl .
\end{align*}
Indeed, we deduce from a change of variable that
\[
\|\phi_T\|_{L_p}^p \lel T^{p-1}\int |\phi(Tx)|^p Tdx
\lel T^{p-1} \|\phi\|_{L_p}^p \pl .
\]
This yields ii). Note that $\phi_T'$ is also disjointly supported, and hence
\[
\| f_T'\|_{L_p}^p \lel \sum_j |\al_j|^p \|\phi_T'\|_{L_p}^p \pl .
\]
Then we note that
\[
\int |\phi'_T(x)|^p dx
\lel T^{2p-1} \int |\phi'(x)|^p dx
\lel T^{2p-1} \|\phi'\|_{L_p}^p \pl .
\]
This implies
\[
\| f_T'\|_{L_p} \lel \Big(\sum_{j=1}^n |\al_j|^p\Big)^{1/p} T^{2-1/p} \|\phi'\|_{L_p}
\lel \frac{\|\phi'\|_{L_p}}{\|\phi\|_{L_p}} T \|f_T\|_{L_p} \pl .
\]
Assertion iii) is a special case.  \qd

The signal model in Lemma~\ref{superposition_model} is given as a superposition of shifts of a given function $\phi$ with finite support and is considered as a generalization of the cardinal B-spline \cite{unser1999splines}. The shifts in Lemma~\ref{superposition_model} are not necessarily on a grid whereas the knots in the cardinal B-spline are integer valued.

Next we illustrate how Corollary~\ref{infshift} implies the RIP on the model in Lemma~\ref{smooth} for two specific choices of $u$ that satisfies \eqref{defwk} for $(w_k)$ given by \eqref{finsuppwk}.

\subsubsection{Deterministic instrument for measurements}

Note that $u: L_2(0,1)\to \ell_2^d$ is described by $h_1,\dots,h_d \in L_2(0,1)$ that satisfy $h_l = u^*(\bm{e}_l)$ for all $l=1,\dots,d$. Suppose that $2N=Ld$ holds for some $L \in \nz$. (This requirement is not a strong restriction in designing a measurement system.) Here we consider a deterministic $u$ with $h_1,\dots,h_d$ given by
\begin{equation}
\label{dethl}
\langle h_l,f\rangle = \sum_{j=1}^L \hat{f}(-N+(l-1)L+j-1), \quad f \in L_2(0,1) \pl ,
\end{equation}
or equivalently
\[
h_l = \sum_{j=1}^L \psi_{-N+(l-1)L+j-1} \pl ,
\]
where $(\psi_k)_{k \in \zz}$ are complex sinusoids defined in \eqref{defpsik}.

Note that $\langle h_l,f\rangle$ is a local average of $(\hat{f}(k))_{-N\le k< N}$ on an interval of size $L$. These measurements are obtained by applying an ideal low-pass filter supported of bandwidth $L$, followed by subsampling by factor $L$ in the Fourier domain. The corresponding weight sequence $w = (w_k)_{k\in\zz}$ satisfies \eqref{finsuppwk}.
For this deterministic $u$, Corollary~\ref{infshift} provides the RIP result shown in the following proposition.

\begin{prop}\label{detv}
Let $u$ be defined as above with $(h_l)_{1\le l\le d}$ satisfying \eqref{dethl}, $t_1,\dots,t_m$ be independent copies of a uniform random variable on $(0,1)$, and $w$ be defined by \eqref{finsuppwk}. Suppose that $2N=Ld$ and $\rho \le N/2$. Then
\[
\mathbb{P}\left(
\sup_{f \in K_{\rho,\gamma}} \Big|\frac{1}{m}\sum_{j=1}^m \|u(\tau_{t_j}f)\|_2^2 -\|f\|_{2,w}^2\Big| \geq \max(\delta,\delta^2)
\right) \leq \zeta
\]
holds provided
\[
m \gl c \delta^{-2} (1+4\rho^2/N^2) \max\Big((2+|\ln\gamma L|)^4 (1+\ln d)^3 (1+\ln m)^3, (2+|\ln\gamma L|) \ln(\zeta^{-1})\Big) \gamma N L \pl .
\]
\end{prop}

\begin{proof}[Proof of Proposition~\ref{detv}]

Lemma~\ref{smooth} implies that every $f \in K_{\rho,\gamma}$ satisfies
\[
\|f\|_{L_q} \leq \sqrt{s} \|f\|_{2,w}
\]
for $s = (1+4\rho^2/N^2) \gamma^{2/q-1}$.

Using the Khintchine inequality and type $2$, we obtain
\begin{align*}
\al_d(u) &= \Big\|\Big(\sum_{l=1}^d |h_l|^2\Big)^{1/2}\Big\|_{L_{q'}}
\le \sqrt{\frac{2}{\pi}} \ez \Big\|\sum_{l=1}^d \eps_l h_l\Big\|_{L_{q'}}
\le \sqrt{\frac{2 q' d}{\pi}} \|h_1\|_{L_{q'}}
\kl C\sqrt{q' d} L^{1-1/q'} \pl ,
\end{align*}
where $1/q + 1/q' = 1$.

Then by Corollary~\ref{infshift}, it suffices to satisfy
\[
m \gl c \delta^{-2} (1+4\rho^2/N^2) \gamma L^2 \max\Big((q')^4 (1+\ln d)^3 (1+\ln m)^3, q' \ln(\zeta^{-1})\Big) d (\gamma^2 L^2)^{-1/q'} \pl .
\]
We chose $q'=\max\{2,|\ln \gamma L|\}$.\qd

\begin{rem}
{\rm
The result in Proposition~\ref{detv} implies that the number of shifts to recover the $\ell_2$-norm of the Fourier series restricted to $[-N,N)\cap\zz$ can be as small as $\gamma N L$ up to a logarithmic factor. The total number of measurements is $\gamma N^2$. In this particular case, if $\gamma = o(1/N)$, then the number of measurements scales sublinearly in the dimension $2N$ of the reconstruction space $\ell_2^{2N}$. Unlike the finite dimensional case, such small $\gamma$ can be still interesting in this infinite dimensional setup. In particular, Proposition~\ref{detv} applies to a sparse function with nonzero measure whereas the existing theory \cite{vetterli2002sampling} only applies to a finite superposition of Dirac's Delta or derivatives, which corresponds to a special case of function with measure zero. We leave it as an open problem to determine the optimal sample rate for this infinite dimensional sparsity model. Nevertheless, as we show in the next section, the number of measurements for a deterministic $u$ can be improved using incoherent measurements with a random instrument $u$.
}
\end{rem}

\subsubsection{Randomized instrument for incoherent measurements}

Next we consider a random instrument $u$ for the measurements. Again we suppose that $2N=Ld$ holds for some $L\in\nz$. Let $(\mathcal{J}_l)_{l=1}^d$ be $d$ non-overlapping ordered sets of size $L$ such that $\cup_{l=1}^d \mathcal{J}_l = [-N,N)\cap\zz$ and $\mathcal{J}_{l} \cap \mathcal{J}_{l'} = \emptyset$ for all $l \neq l'$. Let $k_{l,j}$ denote the $j$th element in $\mathcal{J}_l$, i.e. $\mathcal{J}_l = (k_{l,1},\dots,k_{l,L})$ for $l=1,\dots,d$. Then we consider $u$ with random $h_1,\dots,h_d$ given by
\begin{equation}
\label{randhl}
h_l \lel \sum_{j=1}^L \eps_j \psi_{k_{l,j}}
\end{equation}
with a Rademacher sequence $(\eps_j)_{1\le j\le L}$. Then, by construction, the corresponding weight sequence satisfies \eqref{finsuppwk}.

Randomness in $(h_l)_{1\le l\le d}$ makes the measurements incoherent and reduce the number of translations for the RIP as shown in the following proposition.

\begin{prop}\label{randv}
Let $u$ be defined by random $(h_l)_{1\le l\le d}$ satisfying \eqref{randhl}, $t_1,\dots,t_m$ be independent copies of a uniform random variable on $(0,1)$, and $w$ be defined by \eqref{finsuppwk}. Suppose that $2N=Ld$ and $\rho \le N/2$. Then
\[
\mathbb{P}\left(
\sup_{f \in K_{\rho,\gamma}} \Big|\frac{1}{m}\sum_{j=1}^m \|u(\tau_{t_j}f)\|_2^2 -\|f\|_{2,w}^2\Big| \geq \max(\delta,\delta^2)
\right) \leq 2\zeta
\]
holds provided
\[
m \gl c \delta^{-2} (1+4C^2/N^2) \max\Big((1+|\ln \gamma|)^4 (1+\ln md)^3, (1+|\ln \gamma|) \ln(\zeta^{-1})\Big) \gamma N \pl.
\]
\end{prop}

\begin{proof}[Proof of Proposition~\ref{randv}]
We first compute a tail bound on
\[
\al_d(u) = \Big\|\Big(\sum_{l=1}^d |h_l|^2\Big)^{1/2}\Big\|_{L_{q'}} \pl.
\]

Let $(\eps'_l)_{1\le l\le d}$ be a Rademacher sequence independent of $(\eps_j)_{1\le j\le L}$. Then
\begin{equation}
\label{ubmoment}
\ez_{\eps,\eps'} \Big\|\sum_{l=1}^d \eps'_l h_l \Big\|_{L_{q'}}^{q'} \lel \ez_{\eps,\eps'} \Big\|\sum_{l=1}^d \eps'_l \sum_{j=1}^L \eps_j \psi_{k_{l,j}}\Big\|_{L_{q'}}^{q'} \kl \ez_{\eps''} \Big\|\sum_{k=-N}^{N-1} \eps''_k \psi_k\Big\|_{L_{q'}}^{q'} \pl ,
\end{equation}
where $(\eps''_k)_{-N\le k< N}$ is a Rademacher sequences independent of everything else.
By applying Khintchine's inequality to the upper bound given in \eqref{ubmoment}, we obtain
\[
\Big( \ez_{\eps,\eps'} \Big\|\sum_{l=1}^d \eps'_l h_l \Big\|_{L_{q'}}^{q'} \Big)^{1/q'}
\leq c\sqrt{q' N}
\]
for a numerical constant $c$.

Moreover, by applying Kahane's inequality (see \cite{kwapien1992random}), we obtain
\begin{align*}
\Big(\ez \Big\|\Big(\sum_{l=1}^d |h_l|^2\Big)^{1/2}\Big\|_{L_{q'}}^r\Big)^{1/r}
&\kl \sqrt{\frac{2}{\pi}} \Big(\ez \Big\|\sum_{l=1}^d \eps'_l h_l\Big\|_{L_{q'}}^r\Big)^{1/r} \\
&\kl \sqrt{\frac{2}{\pi}} \max\Big(1, \sqrt{\frac{r-1}{q'-1}}\Big) \Big(\ez \Big\|\sum_{l=1}^d \eps'_l h_l\Big\|_{L_{q'}}^{q'}\Big)^{1/q'} \\
&\le \sqrt{\frac{2}{\pi}} \max\Big(1, \sqrt{\frac{r-1}{q'-1}}\Big) \sqrt{q'}N^{1/2} \\
&\kl c \max\{\sqrt{r},\sqrt{q'}\} N^{1/2} \pl .
\end{align*}
Then, by a consequence of Markov's inequality \cite[Lemma~A.1]{dirksen2015tail},
\begin{equation}
\label{ubheps}
\mathbb{P}\left(
\al_d(u)
\geq \sqrt{e N} (\sqrt{2 \ln(\zeta^{-1})} + \sqrt{q'})
\right) \leq \zeta \pl.
\end{equation}
Under the event when \eqref{ubheps} is satisfied, for the RIP on $K_{\rho,\gamma}$ with probability $1-\zeta$, it suffices to satisfy
\begin{align*}
m \gl c \delta^{-2} (1+4\rho^2/N^2) \gamma L d \gamma^{-2/q'}
\max\Big((q')^3 (1+\ln d)^3 (1+\ln m)^3, \ln(\zeta^{-1})\Big)
(q' + \ln(\zeta^{-1})) \pl .
\end{align*}

The incoherence due to randomness in $(h_l)_{1\le l\le d}$ from the Rademacher sequence $(\eps_j)_{1\le j\le L}$ reduces the number of random translations by factor of $L^{1-2/q'}$.
To optimize $m$, we choose $q'=(1+|\ln \gamma|)$.\qd

\subsection{Preserving valid norms}$\atop$

In general, one cannot recover a sparse signal in an infinite dimensional space from finitely many measurements. For example, when unknown sparse signal $f \in L_2(0,1)$ is supported on a set of nonzero measure, at least a subsequence of the Fourier series at the Landau rate is necessary \cite{landau1967necessary}. Known exceptions include the case where the unknown $f$ corresponds to a point measure, that is a superposition of finitely many Dirac's delta (e.g., \cite{vetterli2002sampling,candes2014towards,tang2013compressed}.

In this section, we demonstrate a similar exception but for $f$ with nonzero measure. As for another application of Corollary~\ref{infshift}, we demonstrate that a set of group structured measurements preserve the weighted norm of a sparse $f \in L_2(0,1)$ according to Definition~\ref{definfsl} up to a small distortion. The recoverability follows as we combine this with the arguments in Section~\ref{sketch}. Specifically, we consider a weighted norm defined by the weight sequence $w = (w_k)_{k\in\zz}$ given as
\begin{equation}
\label{decayingwk}
w_k = \frac{1}{\max(k^2,1)}, \quad \forall k \in \zz \pl .
\end{equation}
Unlike the previous example with a truncated seminorm, $\|\cdot\|_{2,w}$ given by \eqref{decayingwk} is a valid norm. Moreover, it admits a sparsity model on antiderivatives determined by ``smoothness''.

Let $g$ be an antiderivative of $f$ such that $f = g'$ and $\hat{g}(0) = 0$. Such $g$ is uniquely determined and satisfies
\[
\|f\|_{2,w} = \|g'\|_{2,w} = \|g\|_{L_2} \pl.
\]
Therefore there is an one-to-one correspondence between
\[
\{ f \in L_2(0,1) \pl|\pl \|f\|_{L_q} \leq \sqrt{s} \|f\|_{2,w} \}
\]
and
\begin{equation}
\label{set_smooth}
\{ g \in L_2(0,1) \pl|\pl \|g'\|_{L_q} \leq \sqrt{s} \|g\|_2, \pl \hat{g}(0) = 0 \} \pl.
\end{equation}

We will show two examples of group structured measurements $(u(\tau_{t_j}g'))_{j=1}^m$ that preserve the $L_2$-norm of $g$ for all $g$ in the set in \eqref{set_smooth}. Bote that the sparsity level $s$ in \eqref{set_smooth} is clearly interpreted as a measure of smoothness of $g$.

\begin{rem}
\label{withDC}
{\rm
In fact, it is easy to extend the result without requiring $\hat{g}(0) = 0$. We modify the set of smooth signals by
\[
\{ g \in L_2(0,1) \pl|\pl \|g'\|_{L_q} \leq \sqrt{s} (\|g\|_{L_2} - |\hat{g}(0)|^2) \} \pl.
\]
In this case, we need just one more measurement for $\hat{g}(0)$.
}\end{rem}

\subsubsection{Sampling in the time domain}

The first example considers clever scalar-valued measurements. The following theorem shows that one can approximate the $L_2$-norm of a smooth function from evaluation at finitely many points in the time domain.

\begin{theorem}
\label{empirical}
Let $1<q\le 2$ and $0<s<\infty$. Suppose that $t_1,\dots,t_m$ are independent copies of a uniform random variable on $(0,1)$.
Then
\[
\mathbb{P}\left(
\sup_{\begin{subarray}{c}\|g'\|_{L_q}\le \sqrt{s},~ \|g\|_{L_2} = 1 \\ \hat{g}(0)=0\end{subarray}}
\Big|\frac{1}{m} \sum_{j=1}^m |g(t_j)|^2 - \|g\|_{L_2}^2\Big| \geq \delta
\right) \leq \zeta
\]
holds provided
\begin{equation}
\label{m4emp}
m \geq C(q) \delta^{-2} s \max( (q')^3 (1+\ln m)^3, \ln(\zeta^{-1})) \pl .
\end{equation}
Moreover, \eqref{m4emp} also implies
\[
\mathbb{P}\left(
\sup_{\|g'\|_{L_q}\le \sqrt{s}, ~ \|g\|_{L_2}-|\hat{g}(0)|^2 = 1}
\Big|\frac{1}{m} \sum_{j=1}^m (|g(t_j)-\hat{g}(0)|^2+|\hat{g}(0)|^2) - \|g\|_{L_2}^2\Big|
\geq \delta
\right) \leq \zeta \pl.
\]
\end{theorem}

\begin{proof}
Let $f=g'$ for $g$ satisfying $\hat{g}(0)=0$. Then $\|f\|_{2,w} \lel \|g\|_{L_2}$.

We apply Corollary~\ref{infshift} with $d=1$ and $u: L_2(0,1)\to \cz$ given by
\[
u(f) = \langle h, f \rangle \pl , f \in L_2(0,1) \pl ,
\]
where
\[
h=\sum_{j\in\zz\setminus\{0\}} \frac{\psi_j}{j} \pl .
\]

The Hausdorff-Young inequality implies that
\[
\al_d(u) = \|h\|_{L_{q'}} \le \Big\|\Big(\frac{1}{j}\Big)_{j\in\zz\setminus\{0\}}\Big\|_{\ell_q(\zz\setminus\{0\})}
\kl C(q) \pl .
\]

By Corollary~\ref{infshift} \eqref{m4emp} implies
\[
\mathbb{P}\left(
\sup_{\|f\|_{L_q}\le \sqrt{s},~ \|f\|_{2,w} \lel 1}
\Big|\frac{1}{m} \sum_{j=1}^m |u(\tau_{-t_j}f)|^2-\|f\|_{2,w}^2\Big|
\geq \delta
\right) \leq \zeta \pl .
\]

It remains to show $u(\tau_{-t_j}f) = u(\tau_{-t_j}g') = g(t_j)$ for $j=1,\dots,m$.
Note that for every $g$ with $\hat{g}(0) = 0$ satisfies
\[
\sum_{j\in\zz\setminus\{0\}} \frac{\widehat{g'}(j)}{j}
\lel \sum_{j\in\zz\setminus\{0\}} \hat{g}(j) \lel g(0) \pl .
\]
Here the last assertion is obtained by approximation with trigonometric polynomials.
Therefore
\[
\sum_{j\in\zz\setminus\{0\}}
\frac{\widehat{(\tau_{-t_j}(g))'}(j)}{j} \lel (\tau_{-t_j}g)(0) \lel g(t_j) \pl .
\]
The second assertion follows from Remark~\ref{withDC}. \qd

\subsubsection{Taking local measurements in the Fourier domain}

In the previous example, we considered measurements given as time samples. In certain applications, data are acquired sequentially in the Fourier domain. In this situation, before acquiring the full Fourier series, one cannot evaluate time samples. In the second example, we demonstrate a similar result for the measurements obtained as linear combinations of the Fourier series coefficients on finitely supported windows.

Our strategy to get the RIP result on this example consists of the following two steps: First, given a particular partition of $\zz$, we consider infinitely many measurements of random shifts of the signal, where each measurement is taken as the sum over the corresponding block in the partition; Second, we show that due to the fast decay of the weight sequence truncating the measurements to within first few dominant ones does not incur any further significant distortion.

We start with the first step. Note that Corollary~\ref{infshift} does not immediately apply for a map $u:L_2(0,1)\to \ell_2$ with an infinite dimensional range. Instead of working directly on $u$, we approximate $u$ with a sequence of operators $(u_l)_{l\in \zz}$ with finite dimensional ranges and apply Corollary~\ref{infshift} to each $u_l$. Let $(\mathcal{J}_l)_{l \in \zz}$ be a collection of disjoint intervals that partition $\zz$. Then we define $(u_l)_{l\in \zz}$ by
\begin{equation}
\label{defvl}
(u_l(f))(j) =
\begin{cases}
(u(f))(j) & j \in \mathcal{J}_l \pl ,\\
0 & \text{otherwise} \pl.
\end{cases}
\end{equation}
Then the range of each $u_l$ has finite dimension and we also have
\[
\|u(f)\|_2^2 = \sum_{l\in\zz} \|u_l(f)\|_2^2 \pl.
\]

\begin{lemma}
\label{infvl}
Let $t_1,\dots,t_m$ be independent copies of a uniform random variable on $(0,1)$, $(u_l)_{l\in\zz}$ be defined by \eqref{defvl}, $\|\cdot\|_{2,w}$ be a weighted norm defined by \eqref{def2wnorm} with $(w_k)_{k\in\zz}$ given by
\begin{equation}
\label{defwkvl}
w_k = \sum_{l\in\zz} \|u_l(\psi_k)\|_2^2, \quad k \in \zz \pl .
\end{equation}
Suppose that there exists a sequence $(b_l)_{l\in\zz}$ such that
\[
\Big\|\Big(\sum_{j\in \mathcal{J}_l} |u_l^*(\bm{e}_j)|^2\Big)^{1/2}\Big\|_{L_{q'}}\kl b_l \pl , \quad \forall l \in \zz \pl .
\]
Then
\[
\mathbb{P}\left(
\sup_{\|f\|_{L_q} \leq \sqrt{s},~ \|f\|_{2,w} = 1} \Big|\frac{1}{m}\sum_{j=1}^m \sum_{l\in\zz} \|u_l(\tau_{t_j}(f))\|_2^2-\|f\|_{2,w}^2\Big|
\geq \max\{ \delta,\delta^2\} \right) \leq \zeta
\]
holds provided
\[
m \geq
c \delta^{-2} s \max\Big((q')^3 \sum_{l\in\zz} (1+\ln d_l)^3 (1+\ln m)^3 b_l^2, ~ \ln(\zeta^{-1}) \sum_{l\in\zz} b_l^2 \Big) \pl .
\]
\end{lemma}

\begin{proof}
Let $(u_l\tau_{t_j})_{1\le j\le m}:L_q(0,1)\to \ell_{\infty}^m(\ell_2^{d_l})$ denote the composite map such that
\[
(u_l\tau_{t_j})_{1\le j\le m}(f) = (u_l(\tau_{t_j}f))_{1\le j\le m}, \quad f \in L_q(0,1) \pl.
\]
Let $r \in \mathbb{N}$ and $\xi_1,\dots,\xi_m$ be independent copies of $\xi \sim \mathcal{N}(0,1)$.
Conditioned on $t_1,\dots,t_m$, by the triangle inequality together, we have
\begin{align*}
\Big(\ez \sup_{f\in D} \Big|\frac{1}{m}\sum_{j=1}^m \sum_{l\in\zz} \xi_j \|u_l(\tau_{t_j}f)\|_2^2\Big|^r\Big)^{1/r}
\le \sum_{k\in\zz} \Big(\ez \sup_{f\in D} \Big|\frac{1}{m}\sum_{j=1}^m \xi_j \|u_l(\tau_{t_j}f)\|_2^2\Big|^r\Big)^{1/r} \pl.
\end{align*}

By applying \cite[Lemma~2.4]{junge2016ripI} to each summand, we continue as
\begin{align*}
&\Big(\ez \sup_{f\in D} \Big|\frac{1}{m}\sum_{j=1}^m \sum_{l\in\zz} \xi_j \|u_l(\tau_{t_j}f)\|_2^2\Big|^r\Big)^{1/r} \\
&\le C_1 \sqrt{s} \sum_{l\in\zz} \Big(\sup_{f\in D} \sum_{j=1}^m \|u_l(\tau_{t_j}f)\|_2^2\Big)^{1/2} \Big(\E_{2,1}((u_l\tau_{t_j})_{1\le j\le m}) + \sqrt{r} \|(u_l\tau_{t_j})_{1\le j\le m}\|\Big) \\
&\le C_2 \sqrt{s} \Big(\sup_{f\in D} \sum_{j=1}^m \sum_{l\in\zz} \|u_l(\tau_{t_j}f)\|_2^2\Big)^{1/2}
\Big(
\Big(\sum_{l\in\zz} \E_{2,1}^2((u_l\tau_{t_j})_{1\le j\le m}) \Big)^{1/2}
+ \sqrt{r} \Big(\sum_{l\in\zz} \|(u_l\tau_{t_j})_{1\le j\le m}\|^2 \Big)^{1/2}
\Big)
\pl ,
\end{align*}
for numerical constants $C_1$ and $C_2$, where the last step follows from the Cauchy-Schwartz inequality.

Therefore, similarly to the proof of \cite[Proposition~2.6]{junge2016ripI}, it suffices to satisfy
\[
\frac{\sqrt{s}C_3}{\sqrt{m}}
\max\Big(
\Big(\sum_{l\in\zz} \E_{2,1}^2((u_l\tau_{t_j})_{1\le j\le m})\Big)^{1/2}, ~
\Big(\sum_{l\in\zz} \ln(\zeta^{-1}) \|(u_l\tau_{t_j})_{1\le j\le m}\|^2\Big)^{1/2}
\Big) \leq \delta \pl .
\]

Moreover, similarly to the proof of Corollary~\ref{infshift}, we have
\[
\E_{2,1}((u_l\tau_{t_j})_{1\le j\le m}) \leq c_1 (\sqrt{q'})^3 (1+\ln d_k)^{3/2} (1+\ln m)^{3/2} b_l \pl
\]
and
\[
\|(u_l\tau_{t_j})_{1\le j\le m}\| \leq b_l
\]
for all $l\in\zz$.

Applying the upper bounds on the $\E_{2,1}$-norm and the operator norm of $(u_l\tau_{t_j})_{1\le j\le m}$ to the above inequality completes the proof. \qd

Next we consider a specific collection $(u_l)_{l\in\zz}$ such that the corresponding weight sequence $(w_k)_{k\in\zz}$ given by \eqref{defwkvl} satisfies the decaying property in \eqref{decayingwk}.

\begin{exam}\label{tt} {\rm
Let $h_0 = \psi_0$ and
\[
h_l = \sum_{2^{l-2} < |k| \leq 2^{l-1}} \frac{\psi_k}{k}, \quad \forall l \in \nz \pl ,
\]
where $(\psi_k)_{k\in\zz}$ denote complex sinusoids defined in \eqref{defpsik}.
Define a sequence of maps that generate scalar measurements by
\[
u_l(f) =
\begin{cases}
\langle h_l, f \rangle & l \geq 0 \pl , \\
0 & l < 0 \pl .
\end{cases}
\]
Then $(w_k)_{k\in\zz}$ given by \eqref{defwkvl} satisfies \eqref{decayingwk}.

Now we apply Lemma~\ref{infvl} with $\mathcal{J}_l = \{l\}$ for all $l\in\zz$.
In this setup, we have
\[
\Big\|\Big(\sum_{j\in \mathcal{J}_l} |u_l^*(\bm{e}_j)|^2\Big)^{1/2}\Big\|_{L_{q'}}
= \|h_l\|_{L_{q'}} \pl .
\]
Furthermore, by the Hausdorff-Young inequality, we have
Note that
\[
\|h_l\|_{L_{q'}}
= \Big\|\sum_{2^{l-2} < |k| \leq 2^{l-1}} \frac{\psi_k}{k}\Big\|_{L_{q'}}
\kl 2 \Big(\sum_{k \gl 2^{l}} k^{-q}\Big)^{1/q}
\kl 2 C_1(q) 2^{-l/q'} \pl.
\]

Let $g$ be an antiderivative of $f$ such that $f = g'$ and $\hat{g}(0) = 0$. Then $\|f\|_{2,w} = \|g\|_{L_2}$.

Lemma~\ref{infvl} implies
\[
\mathbb{P}\left(
\sup_{\begin{subarray}{c} \|g'\|_{L_q} \leq \sqrt{s}, ~ \|g\|_{L_2} = 1 \\ \hat{g}(0)=0 \end{subarray}} \Big|\frac{1}{m}\sum_{j=1}^m \|u(\tau_{t_j}g')\|_2^2-\|g\|_2^2\Big|
\geq \delta
\right) \leq \zeta
\]
holds provided
\[
m \geq C_3(q) \delta^{-2} s \max( (q')^3 (1+\ln m)^3, \ln(\zeta^{-1})) \pl .
\]
Here $m$ denotes the number of translations for the RIP.

Note that by the chain rule $\tau_{t_j}g'=(\tau_{t_j}g)'$. Moreover, the shift in the time domain becomes a multiplication with a complex exponential in the Fourier domain. Therefore,
\begin{equation}
\label{altrep_fmeas}
u_l(\tau_{t_j}g') = \sum_{2^{l-2} < |k| \leq 2^{l-1}} \widehat{\tau_{t_j}g}(k)
= \sum_{2^{l-2} < |k| \leq 2^{l-1}} e^{-2\pi \mathfrak{i} k t_j} \hat{g}(k) \pl .
\end{equation}
This implies that each scalar measurement is computed as a linear combination of the Fourier series coefficients of $g$ in a given window. This type of measurements are preferred over time samples in certain applications.
}\end{exam}

Example~\label{tt} demonstrated a sufficient number of translates for the RIP. However, there we took infinitely many measurements for each translate. However, when the weight sequence $(w_j)_{j\in\zz\setminus\{0\}}$ is given as in \eqref{decayingwk}, the $L_2$ norm of $f$ that satisfies $\|f\|_{L_q} \leq \sqrt{s} \|f\|_{2,w}$ is highly concentrated on a finite interval in the Fourier domain. Therefore, one can approximate the $L_2$-norm of a smooth function from finitely many measurements, which is stated in the following theorem.

\begin{theorem}
\label{approx_by_finitely_many}
Let $1<q\le 2$, $s>0$, $\mathcal{I}_0 = \{0\}$, and $\mathcal{I}_l=\{ k \pl|\pl 2^{l-2} < |k| \leq 2^{l-1} \}$ for $l\in\nz$. Let $t_1,\dots,t_m$ be independent copies of a uniform random variable on $(0,1)$.
Then there exists a constant $C(q)$ that depends only on $q$ such that
\[
\mathbb{P}\left(
\sup_{\begin{subarray}{c} \|g'\|_{L_q}\le \sqrt{s},~ \|g\|_{L_2} = 1 \\ \hat{g}(0) = 0 \end{subarray}} \Big|\frac{1}{m} \sum_{j=1}^m \sum_{0 \le l\le l_0} \Big|\sum_{k \in \mathcal{I}_l} e^{-2\pi \mathfrak{i} k t_j} \hat{g}(k)\Big|^2-\|g\|_{L_2}^2\Big| \geq \delta
\right) \leq \zeta
\]
provided
\[
l_0 \gl \max\{1,C(q)(1+|\ln(s/\delta)|)\}
\]
and
\[
m \geq C(q) \delta^{-2} s \max( (q')^3 (1+\ln m)^3, \ln(\zeta^{-1})) \pl .
\]
\end{theorem}

\begin{proof} We repeat the arguments in Example \ref{tt} for $\delta/2$.
Let us consider $g\in L_2(0,1)$ such that $\hat{g}(0)=0$ and $f=g'$, then automatically $\hat{f}(0)=0$ and hence
\[
\mathbb{P}\left(
\sup_{\|f\|_{L_q} \le \sqrt{s},~ \|f\|_{2,w}\le 1}
\Big|\frac{1}{m} \sum_{j=1}^m \|u(\tau_{t_j}f)\|^2-\|f\|_{2,w}^2\Big|
\geq \delta/4
\right) \leq \zeta \pl.
\]
Let us recall that
\[
\|f\|_{2,w}^2
\lel \sum_{j\in\zz\setminus\{0\}} k^{-2}|\hat{f}(k)|^2
\lel \sum_{j\in\zz\setminus\{0\}} k^{-2}|\widehat{g'}(k)|^2 \lel \|g\|_{L_2}^2 \pl .
\]
On the other hand for $u_l(f)=\sum_{k\in \mathcal{I}_l} \hat{f}(k)/k$ we have
\begin{align*}
\sum_{l\gl l_0} \|u_l(f)\|_2^2
&= \sum_{l\gl l_0}
\Big|\Big\langle \sum_{k\in \mathcal{I}_l} \frac{\psi_k}{k}, f\Big\rangle\Big|^2 \\
&\le \sum_{l\gl l_0} \|f\|_{L_q}^2 \Big\|\sum_{k\in \mathcal{I}_l} \frac{\psi_k}{k}\Big\|_{L_{q'}}^2 \\
&\le s \sum_{l\gl l_0} \Big(\sum_{k\in \mathcal{I}_l} k^{-q}\Big)^{2/q} \\
&\le C_1(q) s \sum_{l\gl l_0} 2^{-2l/q'}
\kl C_2(q) s 2^{-2l_0/q'} \pl .
\end{align*}
Thus the assertion follows by choosing $l_0$ so that $2^{-2l_0/q'}\kl \frac{\delta}{2C_2(q)s}$ is satisfied. \qd

\begin{rem}
{\rm
Now the number of measurements per translate reduced to a finite number $l_0$ and the total number of scalar measurements is $l_0m$. Note that the oversampling factor $l_0$ compared to the number of translations $m$ is just in the order of $C(q) \ln(s/\delta)$, which is a mild requirement. Here sparsity is implied by a smoothness condition.
}
\end{rem}

\section*{Acknowledgement}
This work was supported in part by NSF grants IIS 14-47879 and DMS 15-01103.
The authors thank Yihong Wu and Yoram Bresler for helpful discussions.


\end{document}